\documentclass[runningheads]{llncs}

\usepackage{eccvabbrv}
\usepackage{tabularx}
\usepackage{graphicx}
\usepackage{booktabs}

\usepackage[accsupp]{axessibility}  

\usepackage{hyperref}

\usepackage{orcidlink}

\usepackage{orcidlink}
\usepackage{colortbl}
\usepackage{algorithm}
\usepackage{algorithmic}
\usepackage{verbatim}
\usepackage{listings}
\usepackage{wrapfig}
\usepackage{subcaption}
\usepackage{multirow}
\usepackage{multicol}
\usepackage{makecell}
\usepackage{booktabs} 

\newcommand{\yty}[1]{\textcolor{black}{#1}}

\begin{document}

\title{SymPoint Revolutionized: Boosting Panoptic Symbol Spotting with Layer Feature Enhancement} 

\titlerunning{SymPoint Revolutionized}

\author{Wenlong Liu\inst{1} \and
Tianyu Yang\inst{1} \and
Qizhi Yu\inst{2} \and
Lei Zhang \inst{1}}

\authorrunning{Wenlong Liu et al.}

\institute{International Digital Economy Academy \and
Vanyi Tech
}

\maketitle

\begin{abstract}

SymPoint \cite{sympoint2024} is an initial attempt that utilizes point set representation to solve the panoptic symbol spotting task on CAD drawing. Despite its considerable success, it overlooks graphical layer information and suffers from prohibitively slow training convergence. To tackle this issue, we introduce SymPoint-V2, a robust and efficient solution featuring novel, streamlined designs that overcome these limitations.  In particular, we first propose a Layer Feature-Enhanced module (LFE) to encode the graphical layer information into the primitive feature, which significantly boosts the performance.  We also design a Position-Guided Training (PGT) method to make it easier to learn, which accelerates the convergence of the model in the early stages and further promotes performance.  Extensive experiments show that our model achieves better performance and faster convergence than its predecessor SymPoint on the public benchmark. Our code and trained models are available at \url{https://github.com/nicehuster/SymPointV2}.

\end{abstract}

\section{Introduction}
\begin{figure}[!t]
  \begin{center}
\includegraphics[width=1.0\linewidth]{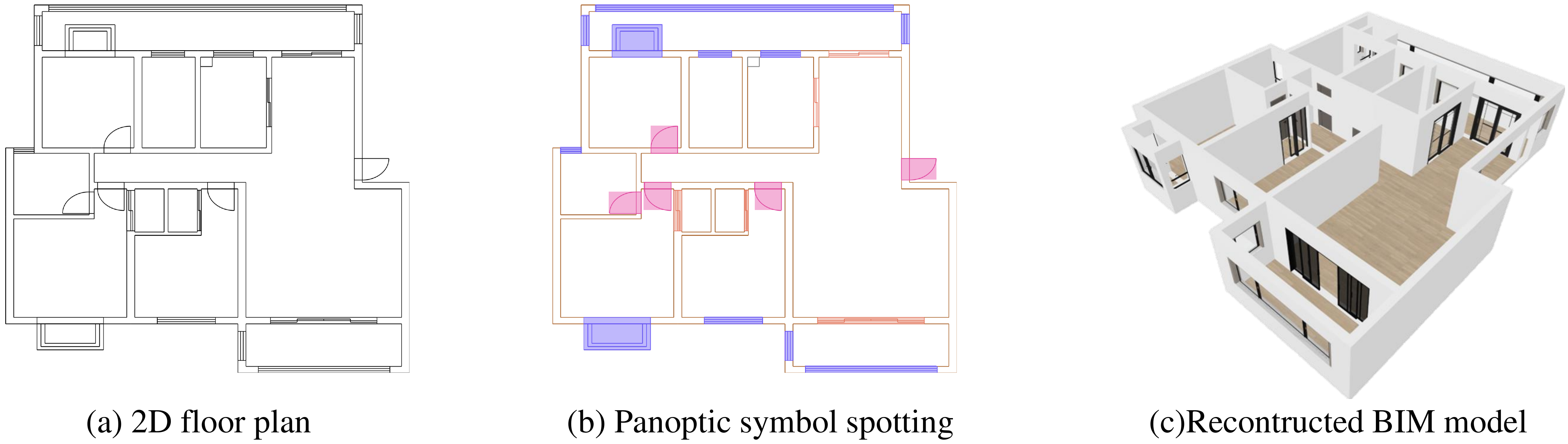}
  \end{center}
  \vspace{-5mm}
\caption{A 2D floorplan (a) and its panoptic symbol spotting results (b), in which the semantics of segments are indicated through different color and instances are highlighted by semi-transparent rectangles. The BIM model (c) with complete semantic and precise geometry can be reconstructed from such an annotated floor plan. We only present the 3D model of windows, doors, and walls for clarity.}
\vspace{-3mm}
\label{bim}
\end{figure}

Symbol spotting is a fundamental task in computer graphics and vision and has a broad range of applications, including document image analysis community\cite{rezvanifar2019symbol} and architecture, engineering and construction (AEC) industries\cite{fan2021floorplancad}. 
In architecture, CAD drawings are instrumental in presenting the exact geometry, detailed semantics, and specialized knowledge relevant to product design, with basic geometric primitives, such as line segments, circles, ellipses, arcs and etc. Spotting and recognizing symbols in CAD drawings is a critical initial step in comprehending their contents, essential for a wide range of practical industrial applications. For example, Building Information Modeling (BIM) is increasingly sought after across various architectural and engineering domains, including pipe arrangement, construction inspection, and equipment maintenance. A CAD drawing typically provides a comprehensive depiction of a storey, presented in an orthogonal top-down view. Therefore, a BIM model can be precisely reconstructed from a group of 2D floor plans with accurate semantic and instance annotations, as shown in Fig~\ref{bim}.

Unlike images that are structured on regular pixel grids, CAD drawings are made up of graphical primitives such as segments, arcs, circles, ellipses, polylines, and others. Spotting each symbol (a set of graphical primitives) within a CAD drawing is challenging due to occlusions, clustering, variations in appearance, and a significant imbalance in the distribution of categories.
Typical approaches\cite{fan2021floorplancad,fan2022cadtransformer,goyal2019bridge,rezvanifar2020symbol,ziran2018object} for tackling the task of panoptic symbol spotting in CAD drawings involve initially converting the CAD drawings into images and then processing it with powerful image-based detection or segmentation methods\cite{ren2015faster,sun2019deep}. Another type of methods \cite{jiang2021recognizing,zheng2022gat,yang2023vectorfloorseg}  uses graph convolutional networks to directly recognize primitives, avoiding the procedure of rastering vector graphics into images.
Recently, SymPoint \cite{sympoint2024} provides a novel insight, which treats CAD drawing as a set of 2D points and applies point cloud segmentation methods to tackle it, leading to impressive results. Its superior performance surpasses all other methods, motivating us to further pursue this avenue of exploration.

\begin{figure*}[t!]
    \centering
    \begin{subfigure}{0.45\textwidth}
        \centering
        \includegraphics[width=\textwidth]{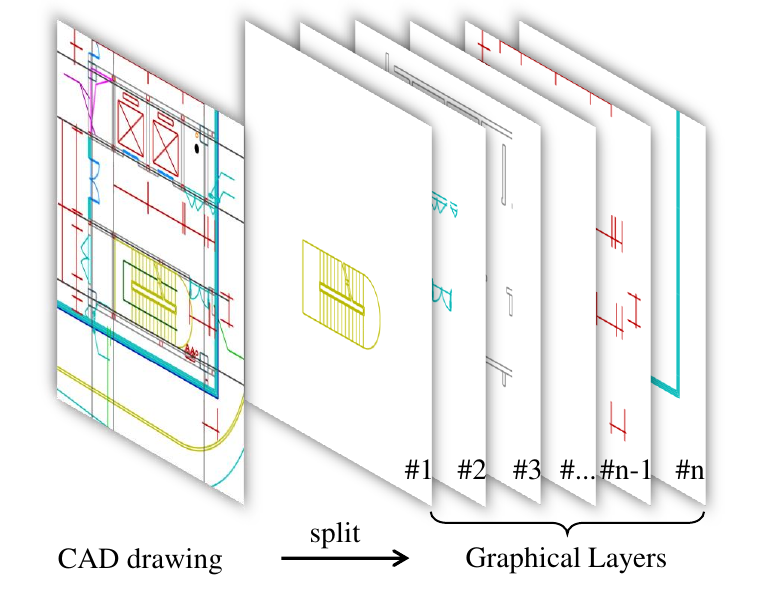}
        \caption{graphical layers}
        \label{fig:layer}
    \end{subfigure}
    \begin{subfigure}{0.45\textwidth}
        \centering
        \includegraphics[width=\textwidth]{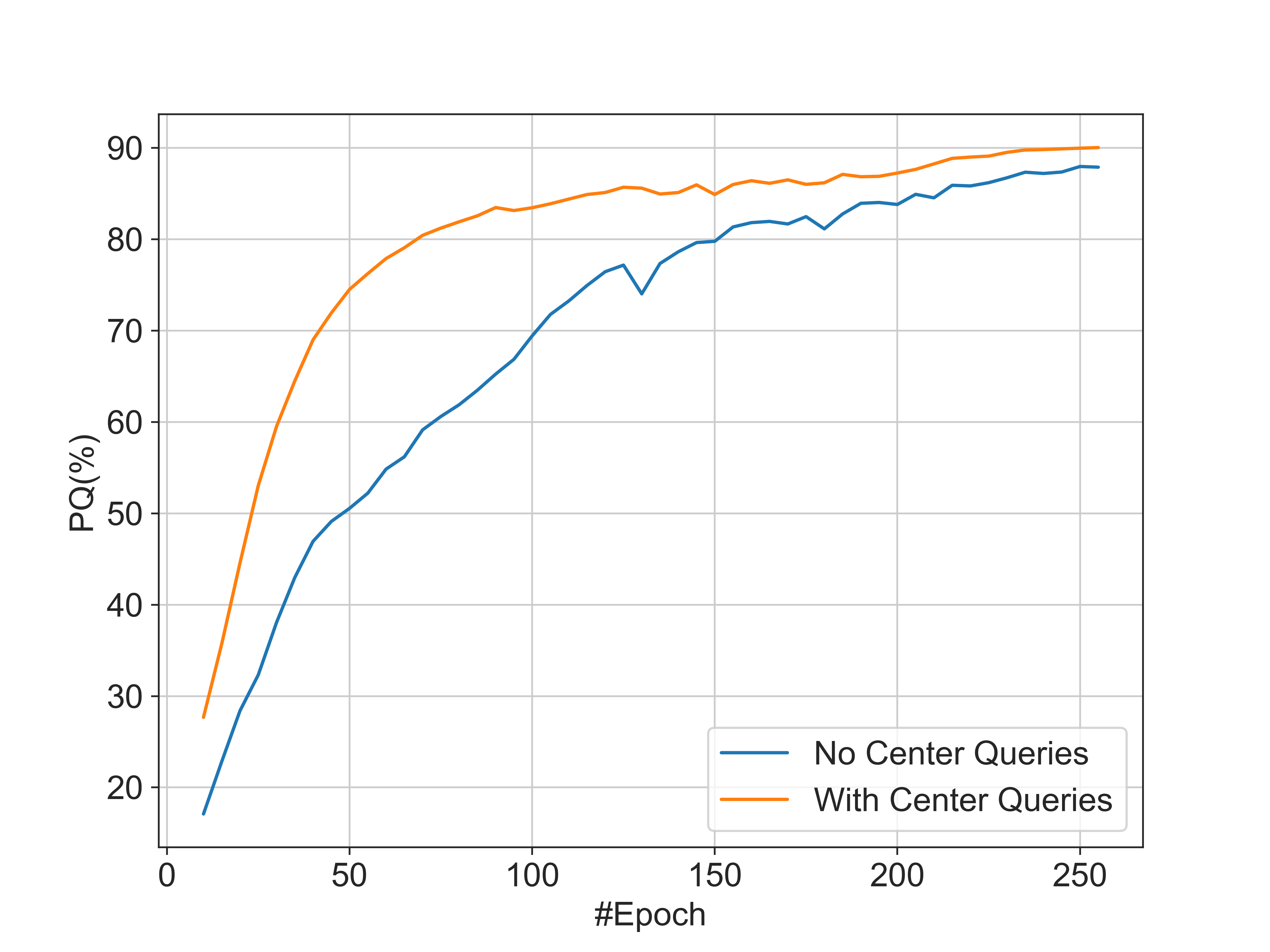}
        \caption{comparision curve.}
        \label{fig:dn_curve}
    \end{subfigure}
    \caption{(a) A CAD drawing is composed of multiple graphical layers. (b) Comparision curves of with and without center queries.}
\end{figure*}

Despite its great success, SymPoint is still an initial attempt which adopts a point-based backbone to extract primitive features and utilizes a transformer decoder to spot and recognize symbols. On the one hand, the former \emph{\textbf{ignores the graphical layer information}} of CAD drawings, which can assign objects of the same or similar types to the same layer and associate them. For example, layers can be created separately for walls, windows, curtains, mobile furniture, fixed furniture, sanitary ware, and etc., to facilitate later drawing management. These layer information is crucial for identifying relationships between primitives. In other words, any CAD drawing can be split into multiple sub-drawings based on graphical layer information, which is crucial for recognizing complex CAD drawings, as shown in \ref{fig:layer}.
On the other hand, current transformer decoder suffers from the issue of \emph{\textbf{slow convergence in the early stages}}. As shown in \ref{fig:dn_curve}, the model (without center queries) manifests slow convergence and lags behind our method by a large margin, particularly in the early stage of training. 

Based on the above observations and analysis, we propose our SymPoint-V2 upon SymPoint \cite{sympoint2024}. we propose two core designs: \yty{Layer Feature-Enhanced (LFE) module and Position-Guided Training (PGT) method. }
LFE aggregates layer information into primitive features, enhancing interaction between primitives in the same layer while PGT adopts a group of additional center queries to guide the training of the transformer decoder, which bypasses bipartite graph matching and directly learns the target, which is crucial in reducing training difficulty and accelerating convergence. 

In conclusion, we propose SymPoint-V2, which improves SymPoint from several perspectives: 
\begin{itemize}
\vspace{-1mm}
	\item{We \yty{proposes a Layer Feature-Enhanced module by} fully utilizing graphical layer information in CAD drawings, which effectively and significantly improves the performance.}
	\item{We \yty{desgin a Position-Guided Training (PGT) method by} constructing a group of center queries for the transformer decoder, which manifests faster convergence and demonstrates higher performance.  }
	\item{Experiments \yty{on public benchmarks} show that our approach achieves a new state-of-the-art result \yty{of 90.1 PQ on FloorplanCAD, surpassing its predecessor SymPoint (83.3 PQ) by a large margin.}}
\end{itemize}

\section{Related Work}
\subsection{Panoptic Symbol Spotting}
Traditional symbol spotting\cite{rezvanifar2019symbol} usually deals with instance symbols representing countable \emph{things}– countable symbols such as windows, tables, sofas, and beds. Following the idea in \cite{kirillov2019panoptic}, \cite{fan2021floorplancad} extended the definition by recognizing semantic of uncountable \emph{stuff} such as wall, railing and parking spot, named it \emph{panoptic symbol spotting}. Therefore, all components in a CAD drawing are covered in one task altogether. Fan et al. \cite{fan2021floorplancad} propose PanCADNet, which adopts Faster-RCNN \cite{ren2015faster} to recognize countable things instances and introduces Graph Convolutional Networks (GCNs) \cite{kipf2016semi} to reason the stuff semantics. Fan et al.\cite{fan2022cadtransformer} propose CADTransformer, 
 instead utilize HRNetV2-W48 \cite{sun2019deep} to tokenize graphical primitives and modify existing ViTs \cite{dosovitskiy2020image} to aggregate graphical primitives' embeddings for the panoptic symbol spotting task. Zheng et al.\cite{zheng2022gat} convert CAD drawing as a graph and utilize Graph Attention Network(GAT) to predict the semantic and instance attributes of every graphical primitive. Besides,  Liu et al.\cite{sympoint2024} pursue a different direction, and propose SymPoint to explore the feasibility of point set representation to tackle panoptic symbol spotting task.

\subsection{Ease Training for DETRs}
Vision transformer is hard to train because globally searching for an object is non-trivial. This phenomenon exists in both detection and segmentation. In detection, DETR\cite{carion2020end} suffers from slow convergence requiring 500 training epochs for convergence. Recently, researchers have dived into the meaning of the learnable queries\cite{liu2022dab,meng2021conditional,wang2022anchor,zhu2020deformable}. They either express the queries as reference points or anchor boxes. \cite{li2022dn,zhang2022dino} proposed to add noised ground truth boxes as positional queries for denoising training and they speed up detection greatly. In segmentation, Mask2Former proposed mask attention which makes training easier and speeds up convergence when compared with MaskFormer. Furthermore, Mask-Piloted (MP) training approach proposed in MP-Former\cite{zhang2023mp} which additionally feeds noised groundtruth masks in masked-attention and trains the model to reconstruct the original ones. Conversely, MAFT\cite{lai2023mask} abandons the mask attention design and resort to an auxiliary center regression task instead.
 
\section{Approach}
\begin{figure}[!ht]
  \begin{center}
\includegraphics[width=1.0\linewidth]{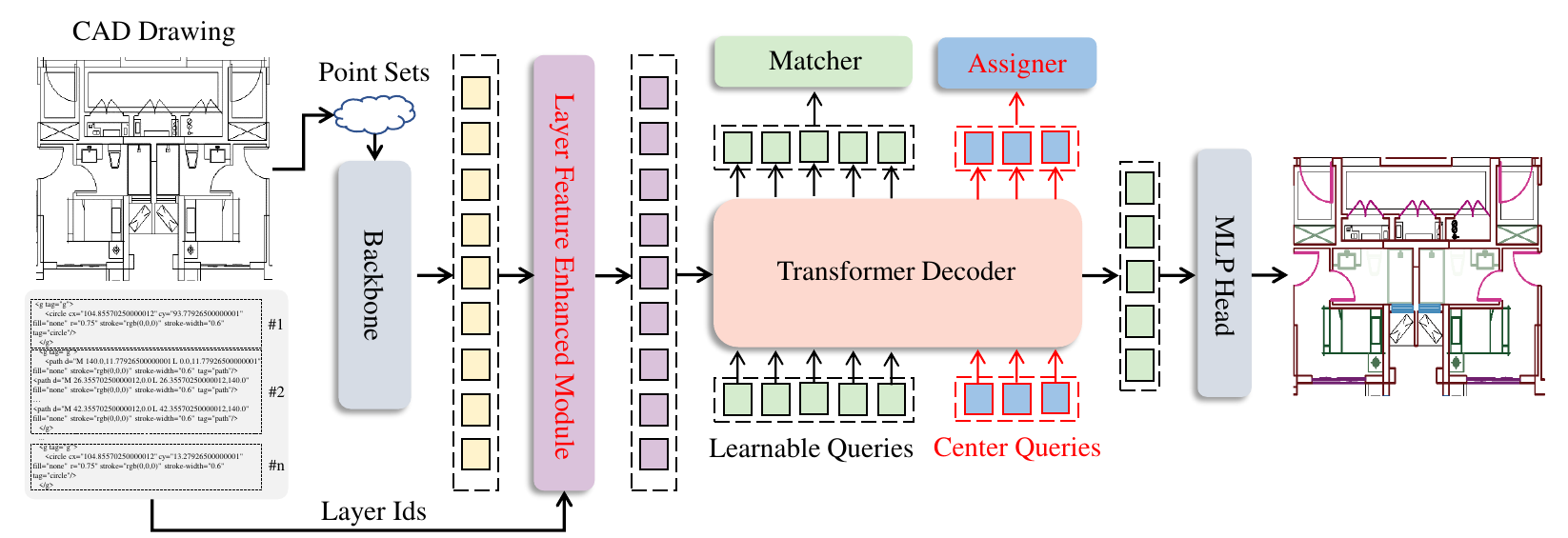}
  \end{center}
  \vspace{-5mm}
  \caption{The overview of our framework.} 
  \label{fig-framework}
\vspace{-3mm}
\end{figure}

We analyze the limitations of SymPoint \cite{sympoint2024} (SPv1) and propose our SymPoint-V2 (SPv2), including two improved modules upon SPv1, As shown in Fig~\ref{fig-framework}. Similar to SPv1, SPv2 receives a CAD drawing and treats it as point sets to represent the graphical primitives, and then the backbone is used to extract primitive features. Subsequently, Layer Feature Enchanced (LFE) module using primitive features and layer information as inputs, integrates layer information to enhance interaction among primitives that are laid out on the same layer. Finally, the enhanced primitive features together with two kinds of query: learnable queries and center queries, are fed into the transformer decoder for query refinement. The first type of query can obtain recognized results through an MLP head, while the second type of query is used to guide the training of the transformer decoder, which bypasses bipartite graph matching and directly assigns ground truth labels to learn the target.

\subsection{Preliminaries}
\subsubsection{Task Formulation.}
Given a CAD drawing represented by a set of graphical primitives $\{p_k\}$, the \textit{panoptic symbol spotting} task requires a map $F_p: p_k\mapsto (l_k, z_k) \in\mathcal{L}\times\mathbf{N}$, where $\mathcal{L}:=\{0,\ldots,L-1\}$ is a set of predetermined set of object classes, and $\mathbf{N}$ is the number of possible instances.
The semantic label set $\mathcal{L}$ can be partitioned into stuff and things subsets, namely $\mathcal{L}=\mathcal{L}^{st}\cup\mathcal{L}^{th}$ and $\mathcal{L}^{st}\cap\mathcal{L}^{th}=\emptyset$. 
We can degrade panoptic symbol spotting to \textit{semantic symbol spotting} task or \textit{instance symbol spotting} task, if we ignore the instance indices or only focus on the thing classes.
\vspace{-5mm}
\subsubsection{SPv1.}
The SPv1\cite{sympoint2024} architecture consists of a backbone, a symbol spotting head, and an MLP head. Firstly, the graphical primitives of CAD drawings are formed as point sets representation $\mathcal P=\{\boldsymbol {p}_k\mid(\boldsymbol {x}_k,\boldsymbol {f}_k)\}$, where $\boldsymbol{x}_k\in\mathbb{R}^{2}$ represents the point position, and $\boldsymbol{f}_k\in\mathbb{R}^{6}$ represents the point features. Secondly, the point sets $\mathcal P$ are fed into the backbone to get the primitive features $\mathcal F \in R^{N \times D}$, where $N$ is the number of feature tokens and $D$ is the feature dimension.
The learnable object queries $\mathcal X$ and the primitive features $\mathcal F$ are fed into the transformer decoder, which refers to symbol spotting head in SPv1\cite{sympoint2024}, resulting in the final object queries, The object queries are parsed to the symbol mask and the classification scores through an MLP head which is mask predicting module in SPv1\cite{sympoint2024}. For each decoder layer $l$, the process of query updating and mask predicting can be formulated as, 
\begin{align}
    X_{l} &= \mathrm{softmax}(A_{l-1} + Q_lK_l^T)V_l + X_{l-1}, \label{eq:query_update}\\
    Y_l &= f_{Y}(X_l),\quad M_l = f_{M}(X_l)F_0^T, \label{eq:prediction}
\end{align}
\label{eq:query_update}
where $X_l \in R^{O \times D}$ is the query features. $O$ is the number of query features. $Q_l = f_Q(X_{l-1})$, $K_l = f_K(F_{r})$ and $V_l = f_V(F_{r})$ are query, key and value features projected by MLP layers. $A_{l-1}$ is the attention mask. The object mask $M_l \in R^{O \times N}$ and its corresponding category $Y_l \in R^{O \times C}$ are obtained by projecting the query features using two MLP layers $f_Y$ and $f_M$, where $C$ is the category number and $N$ is the number of primitives. Meanwhile, the  Attention with Connection Module (ACM) and Contrastive Connection Learning scheme (CCL) are also proposed by SPv1 to effectively utilize connections between primitives.
\vspace{-5mm}
\subsubsection{Baseline.}
We build our baseline upon SPv1, Although connection relationships between primitives are widespread in CAD drawings, their impact on model performance is limited in complex CAD drawings. Therefore, for simplicity, we abandoned ACM and CCL which are proposed by SPv1.

\subsection{Layer Feature Enchanced Module}
\label{sec32}
\begin{figure}[!ht]
  \begin{center}
\includegraphics[width=0.5\linewidth]{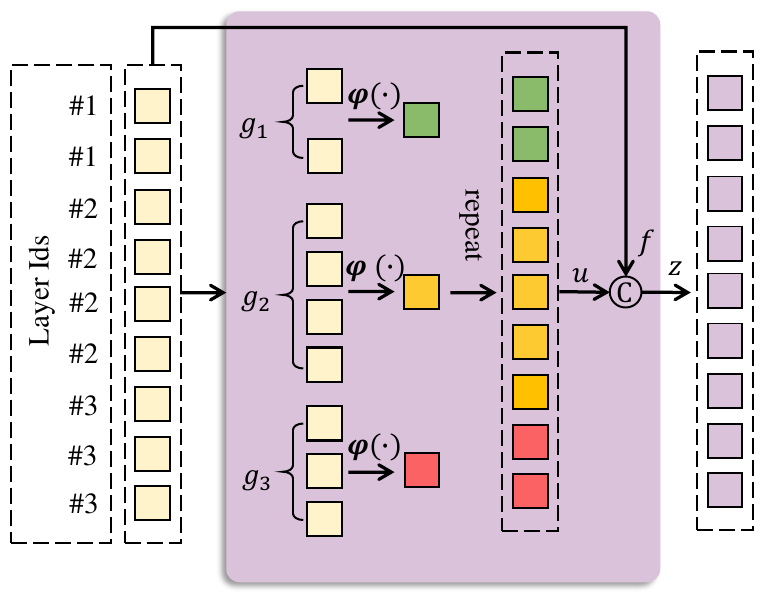}
  \end{center}
  \vspace{-5mm}
  \caption{The framework of our LFE.} 
  \label{lfe-module}
\vspace{-3mm}
\end{figure}
In CAD drawing, graphical layers are tools for effectively organizing and managing design elements. They allow designers to categorize different types of symbols ( walls, windows, curtains, mobile
furniture, fixed furniture, sanitary ware, etc.), facilitating control over visibility, editing, and attribute assignment of these elements. One straightforward idea is to integrate layer information into the process of extracting primitive features in the backbone. But, to be compatible with different point-based backbones, 
We thus propose the Layer Feature-Enhanced (LFE) module and insert it after the backbone. The input of this module is the primitive features $\mathcal F$ and the corresponding layer IDs for each primitive as is shown in Fig.~\ref{lfe-module}. 
This module has two important parts: \emph{pool function $\varphi(\cdot)$} and \emph{fusion function $f(\cdot)$}. The former calculates global layer features, while the latter integrates these global layer features into each primitive feature. 
\vspace{-5mm}
\subsubsection{Pool Function.} Since the layer number can be directly obtained from the CAD drawing, as shown in Fig.~\ref{fig-framework}, after obtaining the primitive features $\mathcal F$ from backbone, it can be divided into $ L$ groups $\mathcal G=\{g_1, g_2, g_3,\ldots, g_L\}$ based on the graphical layer IDs, where $ L$ is total number of graphical layers.

We utilize the pool function for each group of primitive features since the number of primitives laid out on different layers varies greatly. We use a combination of mean pooling $p_1$, max pooling $p_2$, and attention pooling $p_3$ to extract multi-scale global layer features $\mathcal U$. 

\begin{equation}
\mathcal U(g_i)= \varphi(p_1(g_i) \odot p_2(g_i) \odot p_3(g_i)),
g_i\in \mathcal G, i:=\{0,\ldots,L\}
\end{equation} \label{Eq:pool}
where, $\odot$ is concat operation, $\varphi(\cdot)$ is a three-layer MLP.  
\vspace{-5mm}
\subsubsection{Fusion Function.} After extracting global layer features $\mathcal U$, we fuses it and primitive features $\mathcal F$ with broadcast sum or concat. This fusion strategy has the following advantages. (1) \emph{Global-to-Local}. The global layer features with strong layer information can enhance the original primitive features and make global layer information transfer to each primitive feature. (2) \emph{Simple}. This fusion strategy is simple, without introducing extra computational cost. 
In our experiments, we use the concat operation by default.

To integrate layer information to primitive features, we apply LFE module in the mask predicting process. Therefore, Eq.~\ref{eq:prediction} can be reformulated as,
\begin{equation}
Y_l = f_{Y}(X_l),\quad M_l = f_{M}(X_l)\textcolor[rgb]{1,0,0}{f_{LFE}}(F_0)^T,
\label{eq:lfe_prediction}
\end{equation}
where, $f_{LFE}$ is LFE module, and we only applied it on the highest resolution primitives for efficiency.

\subsection{Position-Guided Training}
\label{sec33}
To address the slow convergence problem, inspired by DN-DETR\cite{li2022dn} and MP-Former\cite{zhang2023mp}, we proposed the Position-Guided Training (PGT) method. We construct center queries and along with the learnable queries to feed into the transformer decoder for query refinement. The learnable queries match to GT one by one using bipartite graph matching, while the center queries are assigned to GT to directly learn the target. This training method has the following advantages. (1) \emph{Make learning easier.} The center queries bypass the bipartite graph matching and serve as a shortcut to directly learn mask refinement. By doing so, the transformer decoder learning becomes easier, making bipartite graph matching more stable. (2)\emph{Make learning more stable}. Due to tremendous differences in the distribution of primitives between each graphical layer, the LFE module could easily cause fluctuations in mask cross-entropy loss. The introduction of center queries makes the model converge more stably. 

Our center query consists of two parts: class embedding \textbf{$Q_c$} and positional encoding \textbf{$Q_p$}. The former represents feature information, which can be parsed to the mask/box and the classification scores through an MLP head, while the latter represents positional information, which is the corresponding positional encoding. 
\vspace{-5mm}
\subsubsection{Class Embedding.} We use the class embeddings of ground-truth categories as queries because queries will dot-product with primitive features \yty{to get mask prediction as in Eq. \ref{eq:prediction}} and an intuitive way to distinguish instances/stuff is to use their categories. The class embedding is defined as follows:
\begin{equation}
Q_c = f_{embed}(l)
\end{equation}
where $l$ is ground truth class label and $f_{embed}$ is learnable embedding function.
\vspace{-5mm}
\subsubsection{Positional Encoding.} We take the center of the instance from ground truth and use Fourier positional encodings\cite{tancik2020fourier} to calculate $Q_p$, Since we do not require accurate center coordinates, we perturb the center point to increase diversity, as follows:
\begin{equation}
 Q_p = f_{fourier}(Q_{gt}), Q_{gt} \sim \mathcal N(p_{ct}, {\sigma}^2)
\end{equation}
where $f_{fourier}$ is fourier positional encodings and $p_{ct}$ is instance center. $\mathcal N$ means the Gaussian distribution and $\sigma$ represents deviation. $\sigma =(\epsilon \cdot w, \epsilon \cdot h)$ and $w,h$ is the width and height of instance. $\epsilon$ is the scale factor.

The main difference between our PTG and DN-DETR and MP-Former is that the intrinsicality of DN-DETR and MP-Former are denoising training methods, which feed GT bounding boxes or masks with noises into the transformer decoder and train the model to reconstruct the original boxes or masks. However, our method does not construct any regression task to obtain the accurate object center position, we only construct ground truth center queries to guide the transformer decoder to focus on the position of symbols.
 
\subsection{Training and Inference}

\subsubsection{Training.}
During the training phase, we adopt bipartite matching and set prediction loss to assign ground truth to predictions with the smallest matching cost. 
The overall loss is defined as:
\begin{equation}
\mathcal L= \mathcal L_Q + \mathcal L_{aux}, \mathcal L_Q=\lambda_{bce}L_{bce}+ \lambda_{dice}L_{dice}+\lambda_{cls}L_{cls}
\end{equation}
where $\mathcal L_Q$ is the loss for learnable queries and $\mathcal L_{aux} $ is for center queries. We use the same losses to supervise the center queries. $L_{bce}$ is the binary cross-entropy loss (over the foreground and background of that mask). $L_{dice}$ is the mask Dice loss and $L_{cls}$ is the default cross-entropy loss. The value of $\{\lambda_{bce},\lambda_{dice},\lambda_{cls}\}$ is same as SPv1.
\vspace{-5mm}
\subsubsection{Inference.}
During the test phase, center queries will not be generated. That is, we only parse learnable queries for predicting mask and classification scores by an MLP head. 

\section{Experiments}

\subsection{Experimental Setup}
\subsubsection{Dataset and Metrics.} We conduct our experiments on FloorPlanCAD dataset\cite{fan2021floorplancad}, which has 11,602 CAD drawings of various floor plans with segment-grained panoptic annotation and covering 30 things and 5 stuff classes. We use the panoptic quality (PQ) defined on vector graphics as our main metric to evaluate the performance of panoptic symbol spotting. PQ is defined as the product of segmentation quality (SQ) and recognition quality (RQ), expressed by the formula,
\begin{equation}
PQ = \underbrace{\frac{\sum_{(s_p,s_g)\in TP} \text{IoU}(s_p,s_g)}
{\vert TP \vert}}_{\text{SQ}} \times \underbrace{\frac{\vert TP \vert}{\vert TP \vert + \frac{1}{2} \vert FP \vert + \frac{1}{2} \vert FN \vert}}_{\text{RQ}}
\end{equation}
where a graphical primitive $p = (l, z)$ with a semantic label $l$ and an instance index $z$, $s_p=(l_p, z_p)$ is the predicted symbol. $s_g=(l_g, z_g)$ is the ground truth symbol. A certain predicted symbol is considered as matched if it finds a ground truth symbol, with $l_p=l_g$ and \text{IoU}$(s_p,s_g)>0.5$. The IoU between two primitives is calculated based on arc length $L(\cdot)$,
\begin{equation}
\text{IoU}(s_p, s_g) = \frac{\Sigma_{p_i \in s_p \cap s_g } log(1 + L(p_i)) }
    {\Sigma_{p_j \in s_p \cup s_g } log(1 + L(p_j))}
\end{equation}
The aforementioned three metrics can be adapted for both \emph{thing} and \emph{stuff} categories, represented as ${PQ^{Th}}$, ${PQ^{St}}$, ${RQ^{Th}}$, ${RQ^{St}}$, ${SQ^{Th}}$, ${SQ^{St}}$,respectively. 
\vspace{-5mm}
\subsubsection{Implementation Details.} Our model is trained on 8 NVIDIA Tesla A100 GPUs with a global batch size of 16 for 250 epochs.  Our other basic setup mostly follows the SPv1 framework, except for the following adaptations: 1) The initial learning rate is $2e^{-4}$ and optimizer weight decay is 0.1, while SPv1 is $1e^{-4}$ and 0.001 respectively; 2) We use cosine annealing schedule; 3) We use gradient clipping trick for stable training. As shown in Table \ref{tab-sot}, our baseline method trained for only 250 epochs achieves 82.1 PQ on floorplanCAD while SPv1 trained for 1000 epochs achieves 83.3 PQ.

\begin{figure}[!ht]
  \begin{center}
\includegraphics[width=1.0\linewidth]{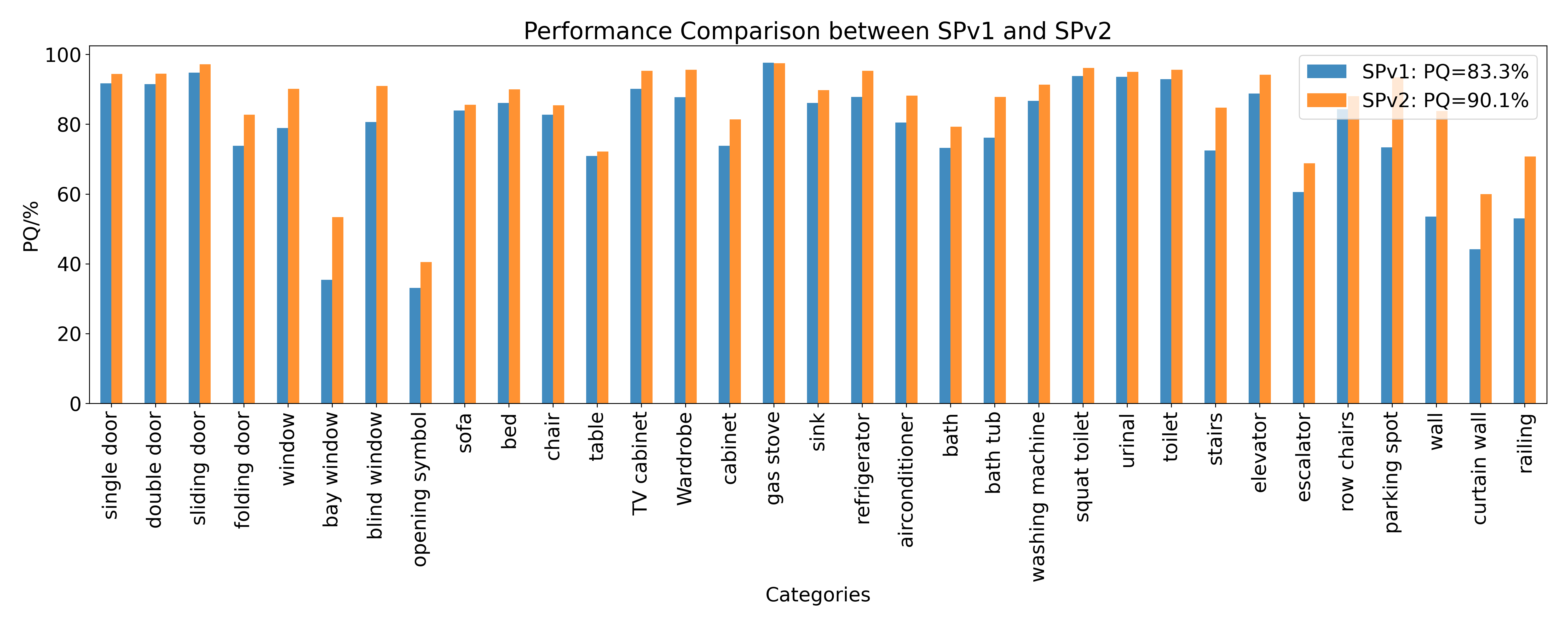}
  \end{center}
  \vspace{-5mm}
  \caption{Performance comparison with SPv1\cite{sympoint2024}, the currently best performing panoptic symbol spotting approach. Per-class PQ results for 35 classes of FloorplanCAD are presented. Note that, we skip the classes that contain less than $1k$ graphical primitives. } 
  \label{comp-spv1}
\vspace{-10mm}
\end{figure}

\subsection{Benchmark Results}
As reported in \cite{fan2021floorplancad,zheng2022gat,fan2022cadtransformer,sympoint2024}, in this section, we also compare our methods with previous works in three tasks: semantic symbol spotting, instance symbol spotting and panoptic symbol spotting. In each benchmark, the \textcolor[rgb]{1,0,0}{\textbf{red bold}} font and the \textcolor[rgb]{0,0,1}{blue} font indicate the best two results. 
\vspace{-5mm}
\subsubsection{Semantic symbol spotting.} We compare our methods with symbol spotting methods\cite{fan2021floorplancad,zheng2022gat,fan2022cadtransformer}. The main test results are summarized in Table~\ref{tab-sem}. Our SPv2 outperforms all existing approaches in the task of semantic symbol spotting. More importantly, compared to SPv1\cite{sympoint2024}, we achieve an absolute improvement of \textbf{2.7\% F1}. and \textbf{2.8\% wF1} respectively.

\begin{table}[t]
\caption{\small \textbf{Semantic Symbol Spotting} comparison results with previous works. wF1: length-weighted F1.}
\label{tab-sem}
\centering
\resizebox{0.95\linewidth}{!}{
\begin{tabular}{c|c|c|c|c|c} 
\Xhline{4\arrayrulewidth}
Methods   &  PanCAD.\cite{fan2021floorplancad} &  CADTrans.\cite{fan2022cadtransformer}  & GAT-CAD.\cite{zheng2022gat} & SPv1\cite{sympoint2024} & \textbf{SPv2(Ours)} \\
\Xhline{2\arrayrulewidth}
F1 & 80.6 &  82.2  & 85.0 & \textcolor[rgb]{0,0,1}{86.8} & \textcolor[rgb]{1,0,0}{\textbf{89.5}} \\ 
wF1 & 79.8 &  80.1  & 82.3 & \textcolor[rgb]{0,0,1}{85.5} & \textcolor[rgb]{1,0,0}{\textbf{88.3}} \\
\Xhline{4\arrayrulewidth}
\end{tabular}}
\end{table}
\subsubsection{Instance symbol spotting.} We additionally conduct comparisons between our method and a range of image detection methods, including FasterRCNN \cite{ren2015faster}, YOLOv3 \cite{redmon2018yolov3}, FCOS \cite{tian2019fcos}, and recent DINO \cite{zhang2022dino}. similar to SPv1\cite{sympoint2024}, We calculate the maximum bounding box of the predicted mask for box AP metric. The
main comparison results are listed in Table~\ref{tab-ins}. Compared to SPv1, we outperform SPv1 by an absolute improvement of \textbf{7.3\% mAP} and \textbf{5.0\% AP50}, respectively. It is worth noting that the additional parameters introduced amount to less than 0.5M, and the inference time has increased by only 29ms.

\begin{table}[t]
\caption{\small \textbf{Instance Symbol Spotting} comparison results with image detection methods.}
\label{tab-ins}
\vspace{-3mm}
\centering
\scriptsize
\resizebox{0.85\linewidth}{!}{\begin{tabular}{lcccc|cc}
\Xhline{3\arrayrulewidth}
\multicolumn{1}{c}{Method}       & Backbone & AP50   & AP75   & mAP  & \#Params & Speed \\ \Xhline{2\arrayrulewidth}
FasterRCNN \cite{ren2015faster}&  R101 &  60.2  &  51.0 & 45.2 &61M  & 59ms \\
YOLOv3 \cite{redmon2018yolov3}&  DarkNet53 &  63.9  &  45.2 & 41.3 &62M  &11ms\\
FCOS \cite{tian2019fcos}&  R101 &  62.4  &  49.1 & 45.3 &51M &57ms\\
DINO \cite{zhang2022dino}& R50 & 64.0 & 54.9 & 47.5 &47M  &42ms\\ 
SPv1\cite{sympoint2024} & PointT\cite{zhao2021point} & \textcolor[rgb]{0,0,1}{66.3} & \textcolor[rgb]{0,0,1}{55.7} & \textcolor[rgb]{0,0,1}{52.8} &35M &66ms \\
\midrule
\textbf{SPv2(ours)} & PointT\cite{zhao2021point} & \textcolor[rgb]{1,0,0}{\textbf{71.3}} & \textcolor[rgb]{1,0,0}{\textbf{60.7}} & \textcolor[rgb]{1,0,0}{\textbf{60.1}} &35M &95ms\\
\Xhline{3\arrayrulewidth}
\end{tabular}}
\label{tab-ins}
\end{table}
\vspace{-5mm}
\subsubsection{Panoptic symbol spotting.} We mainly compare our method with its predecessor SPv1\cite{sympoint2024} , which is the first framework using point sets representation to perform panoptic symbol spotting task. Table~\ref{tab-sot} shows comparison results of panoptic symbol spotting performance. Our method SPv2 surpasses SPv1 by an absolute improvement of \textbf{6.8\% PQ}, \textbf{4.9\% SQ} and \textbf{2.5\% RQ} respectively. Notably, SPv2 greatly outperforms the baseline by an absolute improvement of \textbf{30.5\%} on ${PQ^{St}}$, demonstrating its significant superiority in recognizing \textit{stuff} category. 
Additionally, Fig.~\ref{comp-spv1} presents per-class PQ in the dataset compared to SPv1. Our SPv2 surpasses SPv1 in most classes.

\begin{table*}[!t]
\caption{Panoptic symbol spotting results on FloorplanCAD dataset\cite{fan2021floorplancad}.$^\ddagger$: trained on 1000 epochs.}
\label{tab-sot}
\begin{center}
{\resizebox{1.00\linewidth}{!}{
\begin{tabular}{llcccclcccclccc}
\Xhline{4\arrayrulewidth}

\multirow{2}{*}{Method} & &
\multicolumn{4}{c}{Total} & &
\multicolumn{4}{c}{Thing} & &
\multicolumn{3}{c}{Stuff} 
\\ \cline{3-6} \cline{8-11} \cline{13-15}
& & PQ & SQ & RQ & mIoU & & PQ & SQ & RQ & mAP & & PQ & SQ & RQ
\\ \Xhline{2\arrayrulewidth}

PanCADNet\cite{fan2021floorplancad} & &
59.5 & 82.6 & 66.9 & - & &
65.6 & 86.1 & 76.1 & - & &
58.7 & 81.3 & 72.2
\\ 
CADTransormer\cite{fan2022cadtransformer} & &
68.9 & 88.3 & 73.3 & - & &
78.5 & 94.0 & 83.5 & - & &
58.6 & 81.9 & 71.5
\\ 
GAT-CADNet\cite{zheng2022gat} & &
73.7 & 91.4 & 80.7 & - & &
- & - & - & - & &
- & - & -
\\ 
SPv1$^\ddagger$\cite{sympoint2024} & &
\textcolor[rgb]{0,0,1}{83.3} & \textcolor[rgb]{0,0,1}{91.4} & \textcolor[rgb]{0,0,1}{91.1} & \textcolor[rgb]{0,0,1}{69.7} & &
84.1 & \textcolor[rgb]{0,0,1}{94.7} & 88.8 & 52.8 & &
48.2 & 69.5 & 69.4
\\ 
\hline 
baseline & &
82.1 & 90.8 & 90.4 & 68.7 & &
\textcolor[rgb]{0,0,1}{84.6} & 92.0 & \textcolor[rgb]{0,0,1}{91.9} & \textcolor[rgb]{0,0,1}{52.9} & &
\textcolor[rgb]{0,0,1}{50.3} & \textcolor[rgb]{0,0,1}{70.6} & \textcolor[rgb]{0,0,1}{71.3}
\\ 

\textbf{SPv2(ours)} & &\textcolor[rgb]{1,0,0}{\textbf{90.1}} & \textcolor[rgb]{1,0,0}{\textbf{96.3}} & \textcolor[rgb]{1,0,0}{\textbf{93.6}} & \textcolor[rgb]{1,0,0}{\textbf{74.0}}  & &
\textcolor[rgb]{1,0,0}{\textbf{90.8}} & \textcolor[rgb]{1,0,0}{\textbf{96.6}} & \textcolor[rgb]{1,0,0}{\textbf{94.0}} & \textcolor[rgb]{1,0,0}{\textbf{60.1}} & &
\textcolor[rgb]{1,0,0}{\textbf{80.8}} & \textcolor[rgb]{1,0,0}{\textbf{90.9}} & \textcolor[rgb]{1,0,0}{\textbf{88.9}}
\\ \Xhline{4\arrayrulewidth}
\end{tabular}
}}
\vspace{-5mm}
\end{center}
\end{table*}

\subsection{Qualitative Results}
\begin{figure}[!ht]
  \begin{center}
\includegraphics[width=0.95\linewidth]{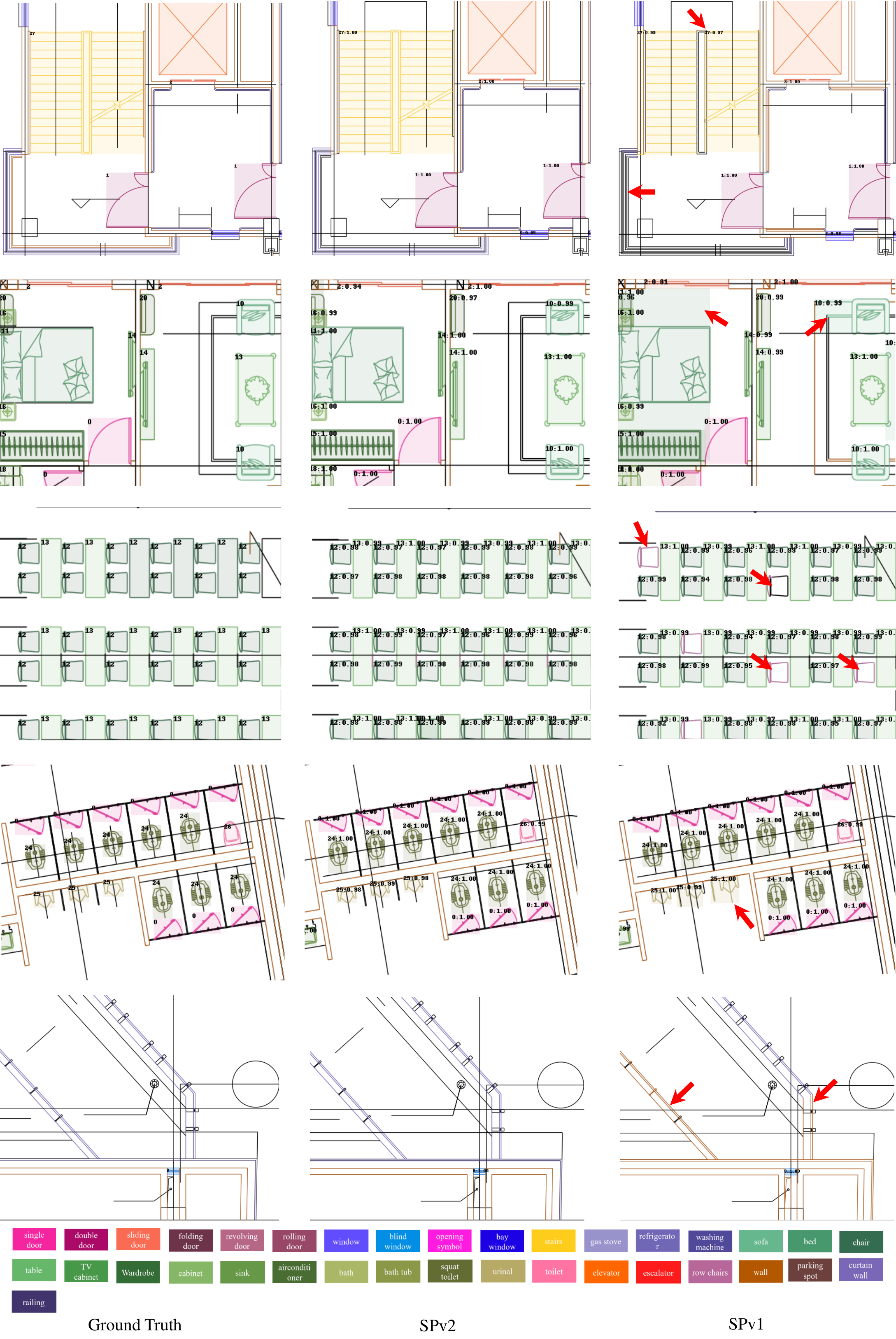}
  \end{center}
  \vspace{-5mm}
  \caption{Visual comparison between SPv1\cite{sympoint2024} and ours. The red arrows highlight the key regions. }
  \label{comp-vis-spv1}
\vspace{-3mm}
\end{figure}

In Fig.~\ref{comp-vis-spv1}, we present qualitative panoptic symbol spotting results on FloorplanCAD as compared to the ground truth masks and those of SPv1\cite{sympoint2024}. The showcased scenes are from the test splits of this dataset, and they are diverse in terms of the type of scenes they exhibit, e.g. residential buildings and core of towers, shopping
malls, and schools. It can be observed that, with our proposed method, more precise instance/stuff masks are obtained as compared to the current state-of-the-art. The highlighted red arrows clearly outline examples where SPv1 predicts wrong instances and merged instances that contain many background primitives, while our method, which effectively utilizes graphical layer information and position guided training method, is able to distinguish between instances and background primitives, and perceive the object position.

\subsection{Ablation Studies}
In this section, we conduct a component-wise analysis to demonstrate the effectiveness of SPv2.
\vspace{-5mm}
\subsubsection{Effects of Components}. We ablate each component that improves the performance of SPv2 in Table~\ref{tab:ablation:components}. Our proposed LFE and PGT  promote the baseline method by absolute \textbf{6.5\% PQ (4.8\% $\mathrm{PQ^{Th}}$,29.4\% $\mathrm{PQ^{St}}$)} and \textbf{2.5\% PQ (3.1\% $\mathrm{PQ^{Th}}$)}, respectively. 
\vspace{-5mm}
\subsubsection{Layer Feature-Enhanced Module.} In section~\ref{sec32}, we design the LFE module to integrate graphical layer information. we make additional analysis on pool types, feature dim of $\varphi(\cdot)$ and multi-level LFE.
\textbf{1) Pool Types.} 
We compare different types of pool function to explore the impact of performance. As shown in Table.~\ref{tab:ablation:pool_type}, our proposed multi-scale global fusion effectively promote of performance.
\textbf{2) Feature Dim of $\varphi(\cdot)$.} 
We ablate the hidden feature dims of MLP $\varphi(\cdot)$ used in Eq. \ref{Eq:pool} to explore its impact on performance. As shown in Table.~\ref{tab:ablation:feat_dim}, the performance can be improved slightly as the number of parameters increases, we select 256 by default for parameter efficiency. 
\textbf{3) Multi-scale LFE.} In section~\ref{sec32}, SPv2 refines learnable queries by iteratively attending to primitive features at different scaled outputs from the backbone. For simplicity, we only apply LFE module to the highest resolution primitive features $\mathcal F_0$ by default. We also provide the result in Table.~\ref{tab:ablation:multi_level} when applying it to multi-scale primitive features $(\mathcal F_0,\mathcal F_1,\mathcal F_2,\mathcal F_3,\mathcal F_4)$. 
It can even lead to improved performance. But it also increases the inference time greatly, reaching 212ms.
\begin{table}[ht]
\caption{Ablation studies on different techniques, pool type, feat dim of $\varphi(\cdot)$, Multi-scale LFE, ablation on ceter query, positional encoding type, and training method.}
\footnotesize
\begin{subtable}{0.6\linewidth}
  \begin{center}
  \scalebox{0.9}{
    \begin{tabular}{ccc|ccc|cc}
    \toprule
    Base & LFE & PGT & $\mathrm{PQ}$ & $\mathrm{PQ^{Th}}$ & $\mathrm{PQ^{St}}$ & Param & Time\\
    \specialrule{0.05em}{3pt}{3pt}
    \checkmark  &      &            &                        82.1 & 84.6 & 50.3 & 35.06M & 66ms\\
    \checkmark  &   \checkmark      &          &             88.6 & 89.4 & 79.7 & 35.14M & 95ms\\
    \checkmark  &      &        \checkmark     &             84.6 & 87.7 & 49.2 & 35.06M & 66ms\\
    \checkmark  &   \checkmark      &      \checkmark &      90.1 & 90.8 & 80.8 & 35.14M & 95ms\\
    \bottomrule
    \end{tabular}}
  \end{center}
  \vspace{-10pt}
  \caption{Ablation studies on different components.}
  \label{tab:ablation:components}
\end{subtable}
\begin{subtable}{0.3\linewidth}
  \begin{center}
  \scalebox{0.9}{
    \begin{tabular}{cc|ccc}
    \toprule
    ClsE. & PosE. & PQ & RQ & SQ\\
    \specialrule{0.05em}{3pt}{3pt}
                &             &88.6 &92.9 & 95.4 \\
    \checkmark  &             &89.2 &93.1 & 95.8 \\
      &  \checkmark           &89.8 & 93.7 & 95.8 \\
    \checkmark & \checkmark    &90.1  &93.6 &96.3 \\
    \bottomrule
    \end{tabular}}
  \end{center}
  \vspace{-10pt}
  \caption{Ablation studies on center query.}
  \label{tab:ablation:position}
\end{subtable}

\begin{subtable}{0.5\linewidth}
  \begin{center}
  \scalebox{0.9}{
    \begin{tabular}{c|p{0.6cm}<{\centering}p{0.6cm}<{\centering}p{0.6cm}<{\centering}p{0.6cm}<{\centering}p{0.6cm}}
    \toprule
     Pool Type  & PQ & RQ & SQ\\
    \specialrule{0.05em}{3pt}{3pt}
      baseline & 82.1 & 90.4 & 90.8 \\
      avepool($p_1$)  & 86.3 & 91.9 & 93.9 \\
      maxpool($p_2$)  & 87.3 & 92.6 & 94.2 \\
      attnpool($p_3$) & 87.9 & 92.8 & 94.7 \\
      concat($p_1 \odot p_2 \odot p_3$)   & 88.6 & 92.9 & 95.4 \\
      \bottomrule
    \end{tabular}}
  \end{center}
  \vspace{-10pt}
  \caption{Pool type}
  \label{tab:ablation:pool_type}
\end{subtable}
\begin{subtable}{0.5\linewidth}
  \begin{center}
  \scalebox{0.9}{
    \begin{tabular}{c|p{0.6cm}<{\centering}p{0.6cm}<{\centering}p{0.6cm}<{\centering}p{0.6cm}<{\centering}p{0.6cm}}
    \toprule
     Feat Dim  & PQ & RQ & SQ\\
    \specialrule{0.05em}{3pt}{3pt}
      128  & 87.1 & 91.3 & 95.4 \\
      256  & 88.6 & 92.9 & 95.4 \\
      512  & 89.1 & 92.9 & 95.9 \\
      1024 & 88.6 & 92.7 & 95.6 \\
      2048 & 88.6 & 92.5 & 95.8 \\
      \bottomrule
    \end{tabular}}
  \end{center}
  \vspace{-10pt}
  \caption{Feature dim of $\varphi(\cdot)$.}
  \label{tab:ablation:feat_dim}
\end{subtable}
\begin{subtable}{0.3\linewidth}
  \begin{center}
  \scalebox{0.9}{
    \begin{tabular}{c|p{0.6cm}<{\centering}p{0.6cm}<{\centering}p{0.6cm}<{\centering}p{0.6cm}<{\centering}p{0.6cm}}
    \toprule
     Multi-scale  & PQ & RQ & SQ\\
    \specialrule{0.05em}{3pt}{3pt}
                   &  90.1 & 93.6 & 96.3\\
      \checkmark   &  90.7 & 94.3 & 96.1\\
      \bottomrule
    \end{tabular}}
  \end{center}
  \vspace{-10pt}
  \caption{Multi-scale LFE}
  \label{tab:ablation:multi_level}
\end{subtable}
\begin{subtable}{0.3\linewidth}
  \begin{center}
  \scalebox{0.9}{
    \begin{tabular}{c|p{0.6cm}<{\centering}p{0.6cm}<{\centering}p{0.6cm}<{\centering}p{0.6cm}<{\centering}p{0.6cm}<{\centering}p{0.6cm}}
    \toprule
    Enc Type & PQ & RQ & SQ\\
    \specialrule{0.05em}{3pt}{3pt}
       Sine    &  89.7 &  93.4 & 96.1 \\
       Fourier  &   90.1 & 94.6 & 95.2 \\
      \bottomrule
    \end{tabular}}
  \end{center}
  \vspace{-10pt}
  \caption{Positional encoding type}
  \label{tab:ablation:pos_type}
\end{subtable}
\begin{subtable}{0.3\linewidth}
  \begin{center}
  \scalebox{0.9}{
    \begin{tabular}{c|p{0.6cm}<{\centering}p{0.6cm}<{\centering}p{0.6cm}<{\centering}p{0.6cm}<{\centering}p{0.6cm}<{\centering}p{0.6cm}}
    \toprule
    Method & PQ & RQ & SQ\\
    \specialrule{0.05em}{3pt}{3pt}
       MPT\cite{zhang2023mp}    &   88.9 & 93.0 & 95.5 \\
       PGT                  &   90.1 & 94.6 & 95.2 \\
      \bottomrule
    \end{tabular}}
  \end{center}
  \vspace{-10pt}
  \caption{Training method}
  \label{tab:ablation:train_method}
\end{subtable}

\end{table}

\begin{figure*}[ht!]
\caption{(a)Comparisions of the recall of instance masks at each trainning epoch. (b) Sensitivity of hyper-parameters on the scale factor.}
\vspace{-10pt}
\centering
\begin{subfigure}{0.3\textwidth}
        \centering
        \includegraphics[width=\textwidth]{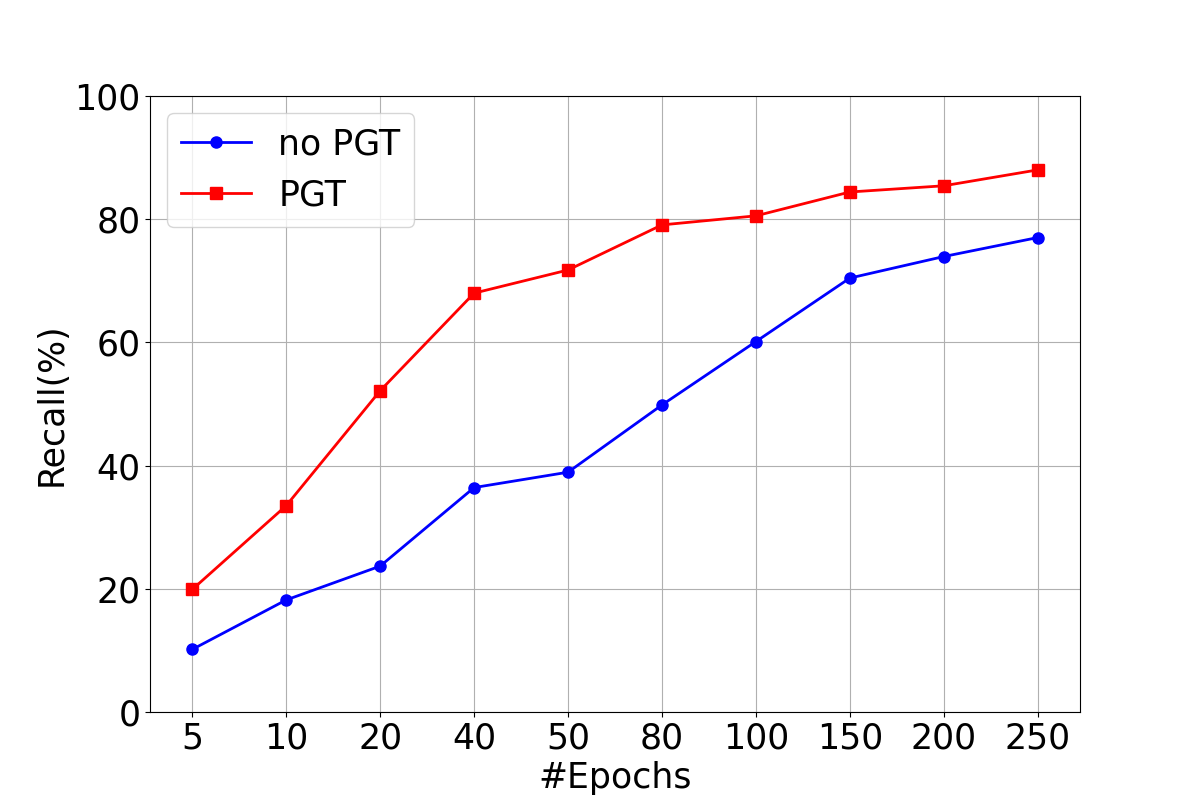}
        \caption{The curve of query recall.}
        \label{recall}
\end{subfigure}
\begin{subfigure}{0.3\textwidth}
        \centering
        \includegraphics[width=\textwidth]{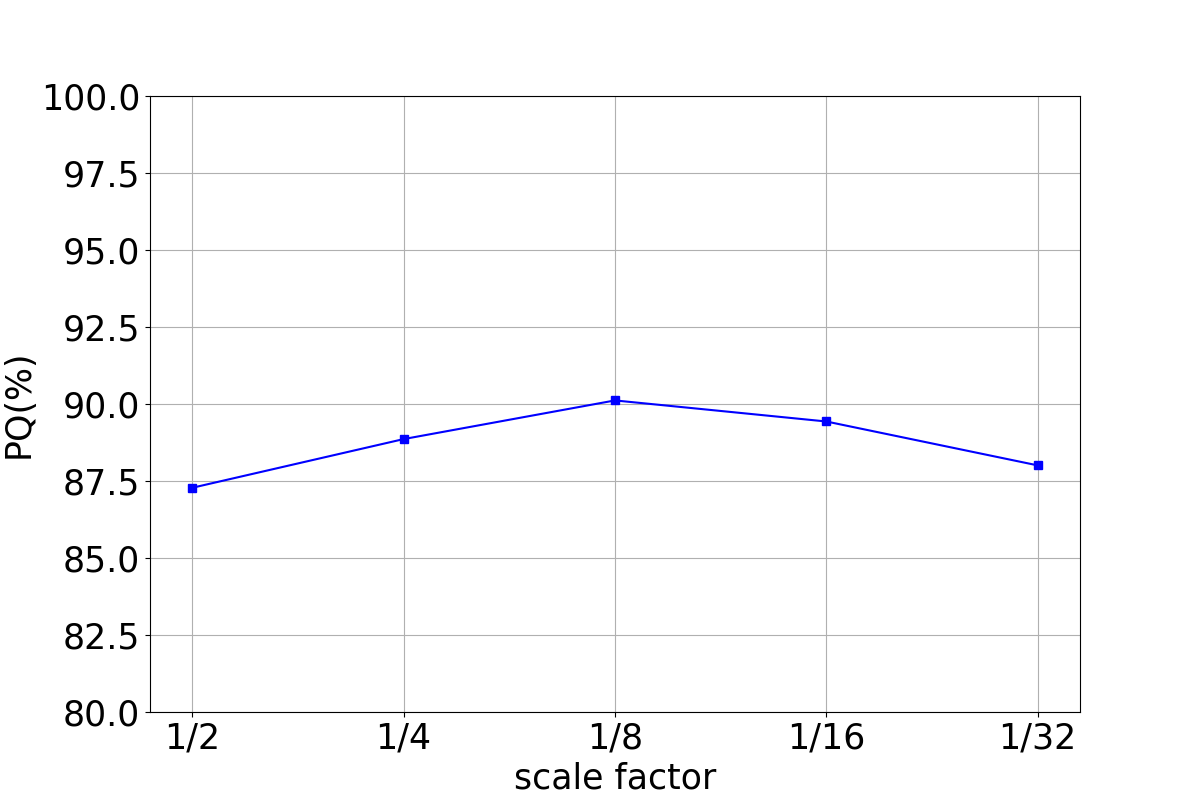}
        \caption{Scale factor.}
        \label{scale_factor}
\end{subfigure}
\begin{subfigure}{0.3\textwidth}
        \centering
        \includegraphics[width=\textwidth]{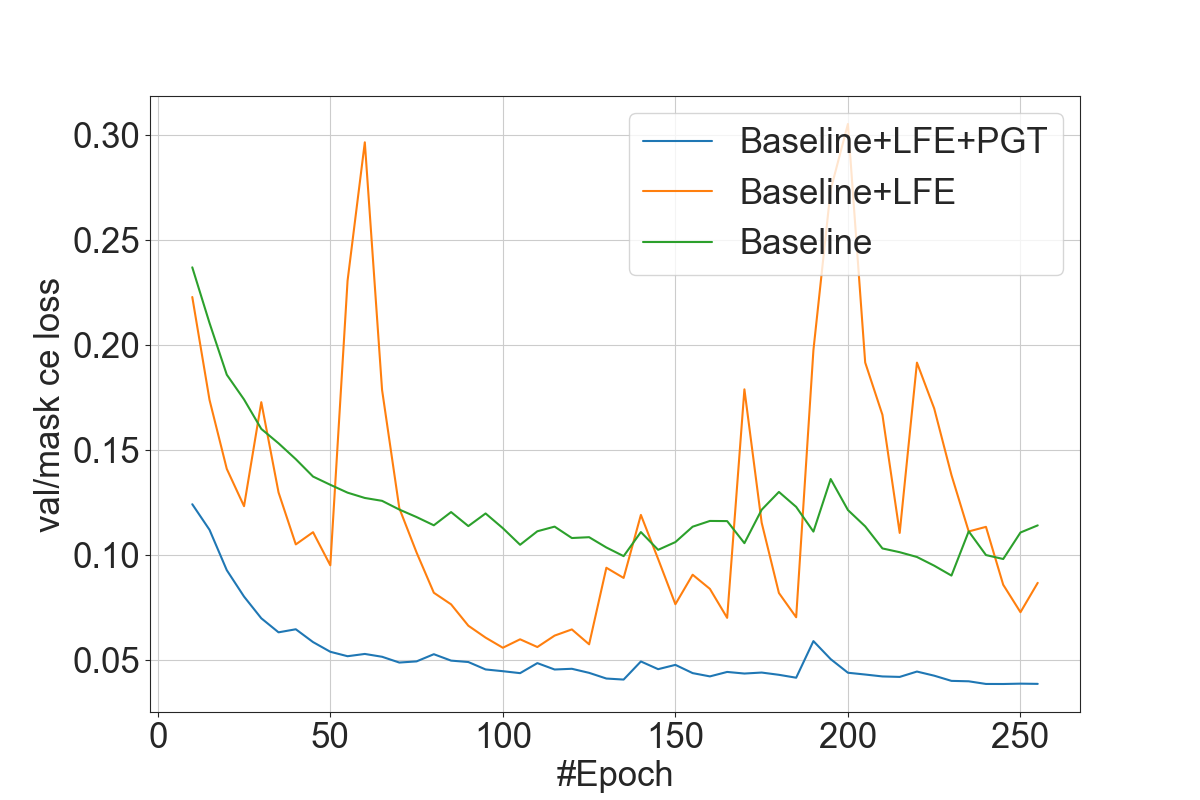}
        \caption{The curve of mask ce loss.}
        \label{mask_ce_loss}
\end{subfigure}
\end{figure*}
\vspace{-5mm}
\subsubsection{Position Guiding Training.} In section~\ref{sec33}, we introduce center queries to implicitly guide model training. As Fig.~\ref{recall} shows, compared to no PGT, PGT can easily capture the objects in a scene with a higher recall in the early stages(before 100 epochs), which is crucial in reducing training difficulty and accelerating convergence. Additionally, we also conduct analysis on ablation of center query, 
sensitivity of scale factor, type of positional encoding, and comparison with MP-Former\cite{zhang2023mp}.
\textbf{1) Ablation of the center query.} We conducted extra ablation analysis on the two parts of the center query: class embedding(ClsE.) and position encoding(PosE.). The results are shown in Table.~\ref{tab:ablation:position}. In line with\cite{zhang2023mp},  using only ClsE. to guide training can also bring an absolute 0.6\% PQ, while using PosE. can get 1.2\% PQ absolute improvement, which confirms the importance of the center position.
\textbf{2) Sensitivity of scale factor.} We perturb the center query to increase diversity by introducing a scale factor to control the degree of distance between the sampling point and the instance center point. The larger the scale factor, the farther away the sampling point is from the instance center point, while the smaller the scale factor, the closer the sampling point is to the instance center point. From Fig.~\ref{scale_factor} we can see that as the sampling point gets closer to the instance center, the performance first increases and then decreases, which means that too far away is not conducive to perceive the position of the object, while too close away reduces diversity and is not conducive to learning.
\textbf{3) Type of positional encodings}. We also experiment with different positional encodings used in the center query. Results can be found in Table.~\ref{tab:ablation:pos_type}. Sine positional encoding can achieve comparable results as Fourier positional encoding, but perhaps the latter is more suitable for our tasks. 
\textbf{4) Stability of mask ce loss.} Due to the huge variety in the distribution of primitives on different graphical layers, directly applying LFE module results in mask ce loss fluctuation during later stages, as the learning rate decreases, as shown in Fig.~\ref{mask_ce_loss}. After introducing position-guided training, convergence can be accelerated in the early stages and stable training can be achieved in the later stages.
\textbf{5) Comparison with MP-Former.} MP-Former\cite{zhang2023mp} feeds noised GT masks and reconstructs the original ones to alleviate inconsistent optimization of mask predictions for image segmentation which bears some similarity with our PGT. We adapt it to our task. As shown in Table.~\ref{tab:ablation:train_method}, our position-guided training method surpasses the mask-piloted training\cite{zhang2023mp} by 2.2\% and 1.6\% in terms of PQ and $\mathrm{PQ^{Th}}$, It shows that our method has a strong modeling ability for instance position.
\begin{figure}[t]
  \begin{center}
\includegraphics[width=0.95\linewidth]{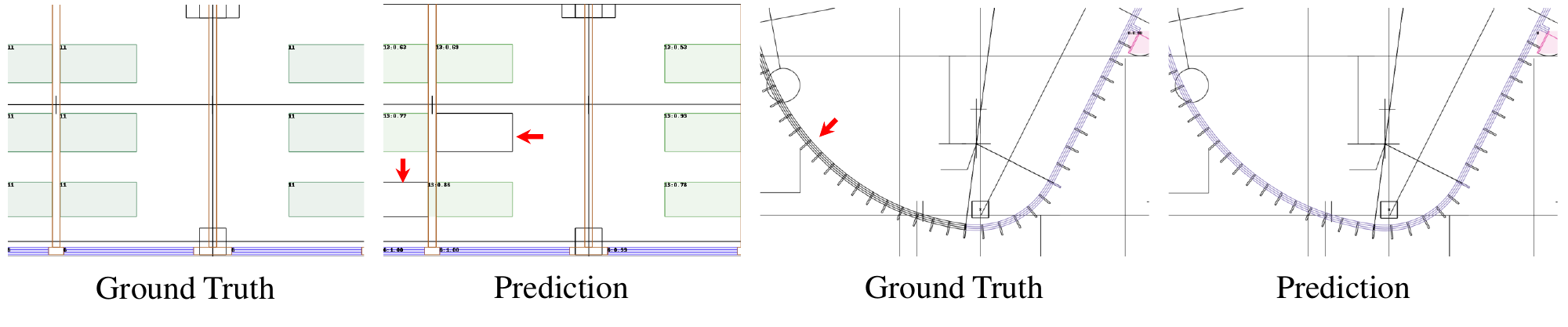}
  \end{center}
  \vspace{-5mm}
  \caption{Two typical failed cases of SPv2. The red arrows highlight the key regions. }
  \label{spv2-failed-case}
\vspace{-3mm}
\end{figure}

\section{Conclusions}
We have presented SymPoint-V2 (SPv2), a simple yet effective approach for panoptic symbol spotting in CAD drawings. Our work makes two non-trivial improvements upon SymPoint-V1\cite{sympoint2024}, including a graphical layer feature-enhanced module to integrate layer information which is laid out in CAD drawing and a position-guided training method. Our SPv2 model achieves new state-of-the-art performance on panoptic symbol spotting benchmarks.

\noindent\textbf{Limitations} Our SymPoint-V2 surpasses existing state-of-the-art methods by a large margin. There are still limitations. Two failed cases are shown in Fig.~\ref{spv2-failed-case}. In some cases, simple symbols may go unrecognized or be incorrectly identified, leading to mislabeling or significant variations in the graphical representation. For example, our model spots most of the quadrilateral tables, but we still missed two tables. Future work would focus on failed cases and improve the robustness of our model.

%
%
\bibliographystyle{splncs04}
\bibliography{main}


\title{Appendix for SymPoint Revolutionized: Boosting Panoptic Symbol Spotting with Layer Feature Enhancement} 

\titlerunning{SymPoint Revolutionized}

\author{Wenlong Liu\inst{1} \and
Tianyu Yang\inst{1} \and
Qizhi Yu\inst{2} \and
Lei Zhang \inst{1}}

\authorrunning{Wenlong Liu et al.}

\institute{International Digital Economy Academy \and
Vanyi Tech
}

\maketitle
\section{PyTorch code for LFE module}
To demonstrate the simplicity of the LFE module,  we can implement our LFE module in several lines of code when the batch size is 1, as summarized in Algorithm~\ref{alg:code}.

\begin{algorithm}[ht]
  \caption{\small PyTorch code for LFE module.}
  \label{alg:code}
  \definecolor{codeblue}{rgb}{0.25,0.6,0.6}
  \definecolor{codekw}{rgb}{0.0, 0.0, 0.0}
  \lstset{
    backgroundcolor=\color{white},
    basicstyle=\ttfamily\selectfont,
    columns=fullflexible,
    numbers=none,
    breaklines=true,
    captionpos=b,
    commentstyle=\fontsize{6.95pt}{6.95pt}\color{codeblue},
    keywordstyle=\fontsize{6.95pt}{6.95pt}\color{codekw},
  }
  \begin{lstlisting}[language=python]
  # F: primitive features tensor with a shape of (N, C)
  # layerids: Layer Ids tensor with a shape of (N, )
  
  # create a blank tensor with the same dimension as F
  new_F = torch.zeros_like(F)
  # do loop processing each layer
  for lid in torch.unique(layerids):
    ind = torch.where(layerids==lid)[0]
    layer_point_feat = element_features[ind]
    avg_pool = torch.mean(layer_point_feat, dim=0)  # mean pool
    max_pool, _ = torch.max(layer_point_feat, dim=0)  # max pool
    # attention pool
    attn_w = F.softmax(self.attn(layer_point_feat), dim=0)
    w_f = torch.mul(layer_point_feat, attn_w.expand_as(layer_point_feat))
    attn_pool = torch.sum(w_f, dim=0)
    # fusion
    layerf = torch.cat((avg_pool,max_pool,attn_pool), dim=0)
    layerf = self.fc1(layerf)
    layerf = F.relu(layerf)
    layerf = self.fc2(layerf)
    layerf = layerf.unsqueeze(0).expand_as(layer_point_feat)
    # concat
    fusion = torch.cat([layer_point_feat, layerf], dim=1)
    output = self.fc3(fusion)
    new_F[ind] = output
  return new_F
  \end{lstlisting}
\end{algorithm}

\section{Additional Quantitative Evaluations}
\vspace{-10pt}
We present a detailed evaluation of panoptic quality(PQ), segmentation quality(SQ), and
recognition quality(RQ) in Tab.~\ref{table:eachclass}. Here, we provide the class-wise evaluations of different setting of our methods.
\begin{table}[ht]
\vspace{-10pt}
\caption{\small \textbf{Quantitative results for panoptic symbol spotting}  of each class. In the test split, some classes have a limited number of instances, resulting in zeros and notably low values in the results.}
\tabcolsep=0.2cm
\centering
\resizebox{\linewidth}{!}{
\begin{tabular}{c|ccc|ccc|ccc|ccc|ccc} 
\toprule
\textbf{Class} & \multicolumn{3}{c}{\textbf{SPv2}}  & \multicolumn{3}{c}{\textbf{Baseline+PGT}}  & \multicolumn{3}{c}{\textbf{Baseline+LFE}}  & \multicolumn{3}{c}{\textbf{Baseline}}  & \multicolumn{3}{c}{\textbf{SPv1}} \\
&PQ   &RQ  &SQ &PQ   &RQ  &SQ &PQ   &RQ  &SQ &PQ   &RQ  &SQ  &PQ   &RQ  &SQ\\  
\bottomrule
single door & 94.4 & 97.1 & 97.3  & 91.6 & 95.9 & 95.5  & 93.4 & 96.3 & 97.1 &  90.5 & 95.1 & 95.1 & 91.7 & 96.0 & 95.5  \\
double door & 94.5 & 97.3 & 97.1  & 91.4 & 96.3 & 94.9 & 93.9 & 96.8 & 97.0 & 90.0 & 95.3 & 94.4 & 91.5 & 96.6 & 94.7  \\
sliding door & 97.2 & 97.9 & 99.3  & 94.6 & 97.5 & 97.0  & 96.8 & 97.6 & 99.2  & 93.8 & 97.5 & 96.2 & 94.8 & 97.7 & 97.0  \\
folding door & 82.7 & 90.0 & 91.9  & 64.6 & 69.8 & 92.6 & 81.7 & 87.2 & 93.7  & 70.3 & 79.1 & 88.9  & 73.8 & 87.0 & 84.8 \\
revolving door & 0.0 & 0.0 & 0.0  & 0.0 & 0.0 &  0.0  & 0.0 & 0.0 &  0.0  & 0.0 & 0.0 &  0.0 & 0.0 & 0.0 & 0.0  \\
rolling door & 0.0 & 0.0 & 0.0  & 0.0 & 0.0 &  0.0  & 0.0 & 0.0 &  0.0  & 0.0 & 0.0 &  0.0 & 0.0 & 0.0 & 0.0  \\
window & 90.1 & 93.1 & 96.8 & 79.3 & 90.7 & 87.4  & 89.4 & 92.7 & 96.4  & 77.8 & 89.6 & 86.8  & 78.9 & 90.4 & 87.3 \\
bay window & 54.1 & 55.1 & 98.3  & 25.2 & 34.2 & 73.8  & 39.8 & 41.4 & 96.2  & 19.4 & 27.1 & 71.5  & 35.4 & 42.3 & 83.6 \\
blind window & 91.0 & 92.3 & 98.5  & 79.6 & 90.8 & 87.7  & 86.1 & 89.1 & 96.6  & 77.6 & 89.6 & 86.6  & 80.6 & 92.1 & 87.5  \\
opening symbol & 40.7 & 51.6 & 78.9  & 48.2 & 61.0 & 79.0 & 32.6 & 42.6 & 76.5  & 35.1 & 45.1 & 77.9   & 33.1 & 40.9 & 80.7 \\
sofa & 85.4 & 89.2 & 95.8  & 84.6 & 90.0 & 94.0  & 83.7 & 88.3 & 94.7  & 82.4 & 87.8 & 93.8  & 83.9 & 88.8 & 94.5 \\
bed & 90.0 & 97.5 & 92.4  & 79.5 & 90.1 & 88.3  & 86.6 & 95.1 & 91.1  & 76.9 & 87.9 & 87.5  & 86.1 & 95.9 & 89.8  \\
chair & 85.5 & 89.7 & 95.3  & 84.6 & 89.6 & 94.4  & 86.3 & 91.0 & 94.9  & 85.7 & 90.7 & 94.5  & 82.7 & 88.9 & 93.1\\
table & 72.1 & 79.7 & 90.5  & 71.5 & 80.0 & 89.4  & 70.2 & 79.3 & 88.4  & 70.9 & 81.1 & 87.3 & 70.9 & 79.1 & 89.6\\
TV cabinet & 95.5 & 97.7 & 97.8  &  92.3 & 96.9 & 95.2 & 92.6 & 96.5 & 96.0  & 87.1 & 95.0 & 91.6 & 90.1 & 97.0 & 92.9  \\
Wardrobe & 95.6 & 97.7 & 97.9  &  86.7 & 97.0 & 89.4 & 94.2 & 96.5 & 97.6 & 85.3 & 96.3 & 88.6 & 87.7 & 96.4 & 90.9  \\
cabinet & 81.3 & 89.7 & 90.6  & 73.5 & 86.5 & 85.0  & 78.9 & 87.2 & 90.5  & 72.8 & 85.9 & 84.8 & 73.8 & 86.2 & 85.6 \\
gas stove & 96.4 & 97.0 & 99.4  & 95.4 & 96.1 & 99.3  & 97.4 & 98.9 & 98.5  & 97.0 & 98.9 & 98.1 & 97.6 & 98.9 & 98.7  \\
sink & 89.8 & 94.2 & 95.3  &  87.2 & 93.2 & 93.6 & 88.6 & 94.1 & 94.1 & 85.5 & 92.7 & 92.2 & 86.1 & 92.9 & 92.7 \\
refrigerator & 95.3 & 96.3 & 98.9  & 94.4 & 95.9 & 98.4 & 88.7 & 96.0 & 92.3  &  87.0 & 95.4 & 91.2 & 87.8 & 95.7 & 91.8  \\
airconditioner & 88.2 & 89.1 & 98.9  & 84.2 & 88.3 & 95.4 & 88.1 & 89.2 & 98.8  & 83.4 & 87.9 & 94.9 & 80.5 & 84.4 & 95.4 \\
bath & 79.2 & 86.2 & 91.9  & 73.0 & 86.2 & 84.7 & 78.4 & 86.6 & 90.5  & 71.1 & 85.5 & 83.1 & 73.2 & 85.0 & 86.1 \\
bath tub & 87.8 & 91.4 & 96.0  &  83.5 & 90.6 & 92.2 & 86.0 & 94.2 & 91.3  & 74.3 & 90.3 & 82.4 & 76.1 & 91.4 & 83.2  \\
washing machine & 91.4 & 93.6 & 97.6  & 87.3 & 93.5 & 93.3  & 89.1 & 92.3 & 96.5  & 84.3 & 92.6 & 91.0 & 86.7 & 93.8 & 92.5 \\
urinal & 95.0 & 95.6 & 99.4 & 93.2 & 95.7 & 97.4  & 94.4 & 95.6 & 98.8  & 91.6 & 95.6 & 95.8 & 93.8 & 96.7 & 96.9  \\
squat toilet & 96.1 & 97.1 & 99.0  & 94.2 & 97.1 & 97.0  & 95.8 & 97.1 & 98.6  & 91.5 & 95.7 & 95.7 & 93.6 & 97.5 & 96.1 \\
toilet & 95.6 & 97.5 & 98.1 & 93.8 & 97.0 & 96.7  & 93.6 & 97.2 & 96.3  & 92.0 & 96.9 & 95.0 & 92.9 & 97.2 & 95.6 \\
stairs & 84.8 & 89.8 & 94.4  & 76.9 & 88.6 & 86.8  & 82.8 & 89.3 & 92.7  & 72.5 & 85.5 & 84.9 & 72.5 & 85.3 & 85.0 \\
elevator & 94.2 & 96.4 & 97.7 &  91.8 & 95.9 & 95.7  & 93.4 & 96.8 & 96.6  & 90.8 & 96.0 & 94.6 & 88.8 & 94.4 & 94.1 \\
escalator & 68.7 & 80.7 & 85.2 & 60.7 & 77.6 & 78.3  & 64.8 & 77.7 & 83.4 & 51.5 & 65.4 & 78.8 & 60.6 & 75.6 & 80.2 \\
row chairs & 88.0 & 92.4 & 95.3 & 85.7 & 90.8 & 94.4  & 84.6 & 89.4 & 94.7  & 84.5 & 89.2 & 94.7 & 84.3 & 89.2 & 94.5 \\
parking spot & 93.6 & 95.6 & 97.9  & 80.1 & 92.3 & 86.8  & 87.5 & 89.7 & 97.6  & 71.2 & 85.1 & 83.7 & 73.4 & 86.7 & 84.7 \\
wall & 83.7 & 92.5 & 90.6  & 54.6 & 79.1 & 69.0  & 82.6 & 91.2 & 90.5  & 50.8 & 75.3 & 67.4 & 53.5 & 77.5 & 69.0  \\
curtain wall & 60.0 & 70.1 & 85.6  & 41.8 & 58.2 & 71.7  & 57.8 & 68.1 & 84.9 & 39.8 & 53.8 & 74.0  & 44.2 & 60.2 & 73.5 \\
railing & 70.7 & 77.0 & 91.8 & 42.3 & 53.5 & 79.0 &  64.6 & 70.8 & 91.2  & 37.7 & 48.2 & 78.1 & 53.0 & 66.3 & 80.0 \\
\bottomrule
total  & 90.1& 93.6 & 96.3  & 84.6 & 91.7 & 92.2  & 88.6 & 92.9 & 95.4  &  82.1 & 90.4 & 90.8 &83.3 &91.1 &91.4 \\
\bottomrule
\end{tabular}
}
\label{table:eachclass}
\end{table}

\section{Additional Qualitative Evaluations}
\begin{figure*}[ht!]
    \centering
    \begin{subfigure}{0.45\textwidth}
    \centering
    \includegraphics[width=\textwidth]{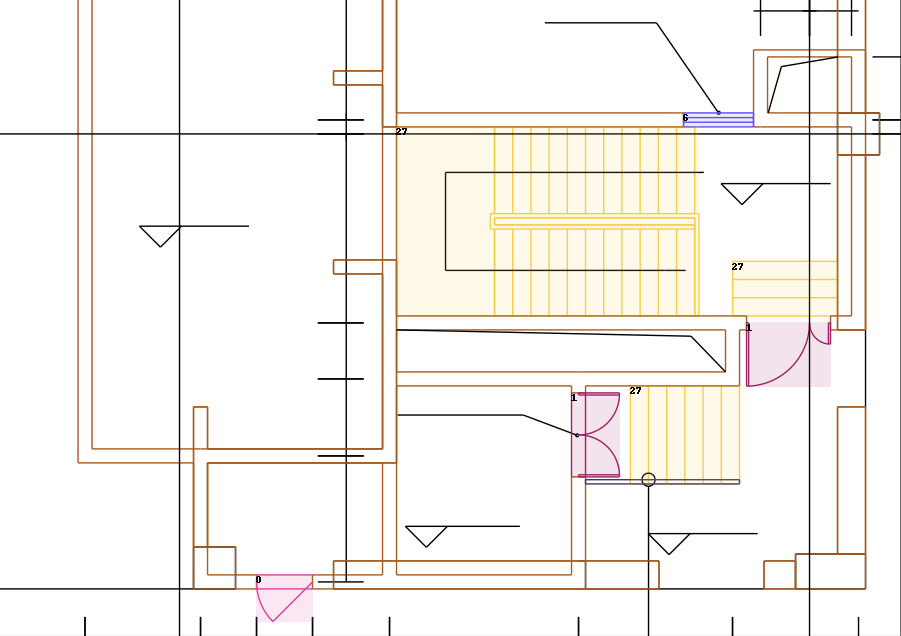}  
    \end{subfigure}
    \begin{subfigure}{0.45\textwidth}
    \centering
    \includegraphics[width=\textwidth]{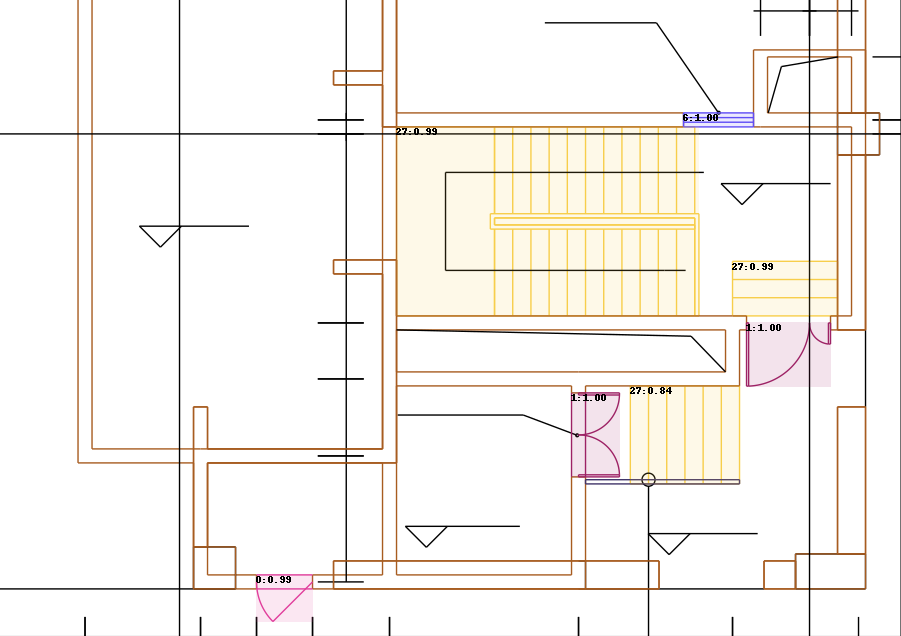}
    \end{subfigure}
    \begin{subfigure}{0.45\textwidth}
    \centering
    \includegraphics[width=\textwidth]{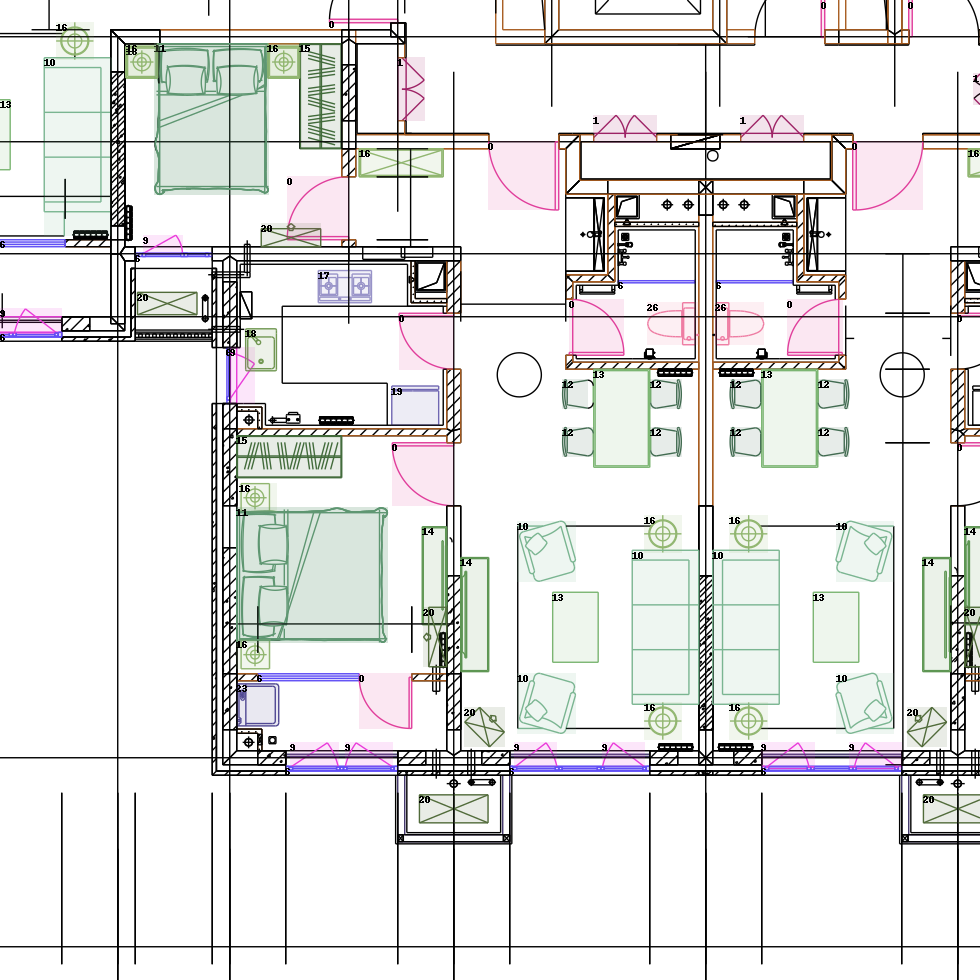}  
    \end{subfigure}
    \begin{subfigure}{0.45\textwidth}
    \centering
    \includegraphics[width=\textwidth]{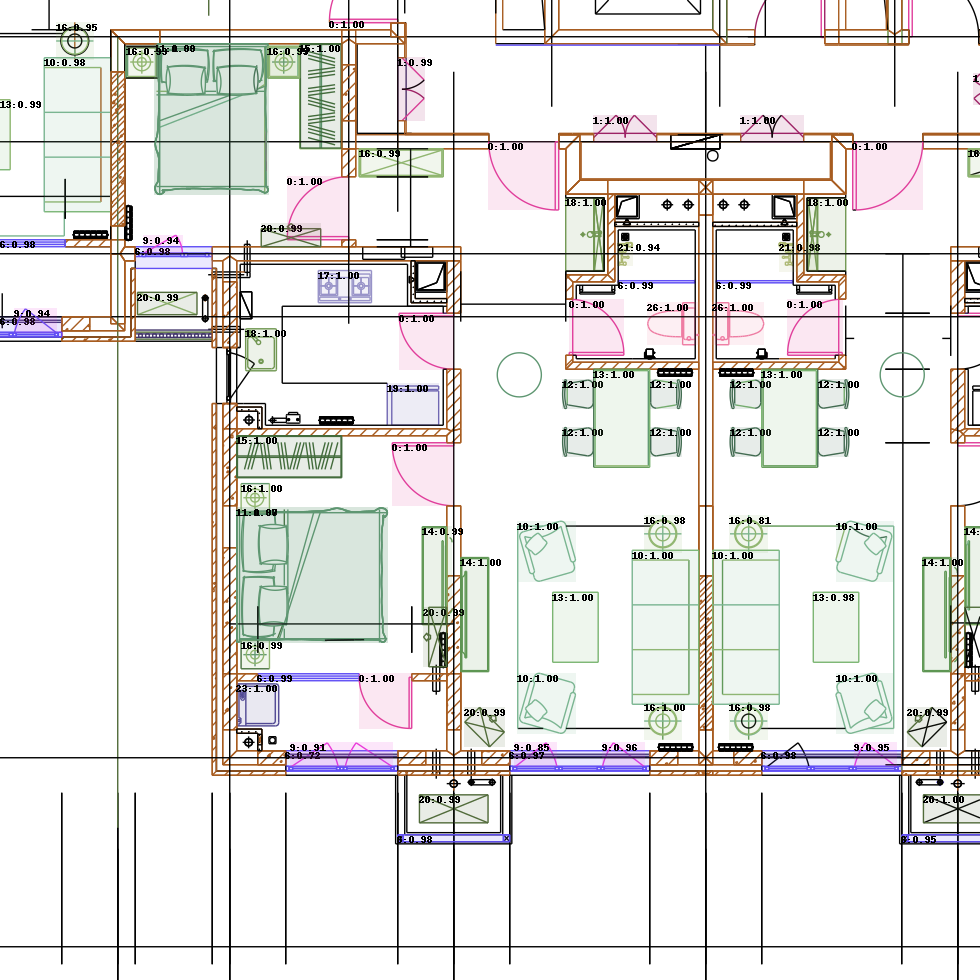}
    \end{subfigure}
    \begin{subfigure}{0.45\textwidth}
    \centering
    \includegraphics[width=\textwidth]{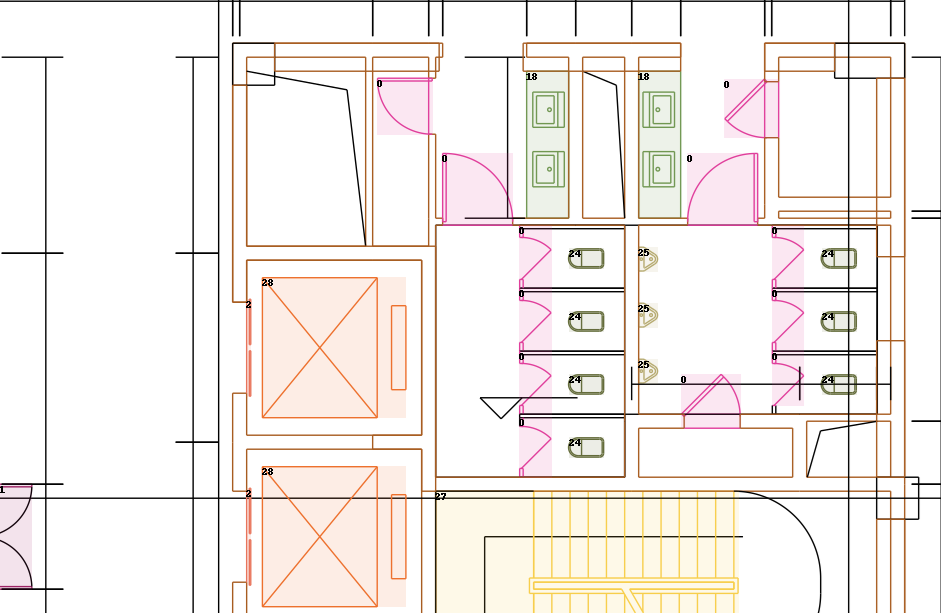}  
    \caption{}{Ground Truth}
    \end{subfigure}
    \begin{subfigure}{0.45\textwidth}
    \centering
    \includegraphics[width=\textwidth]{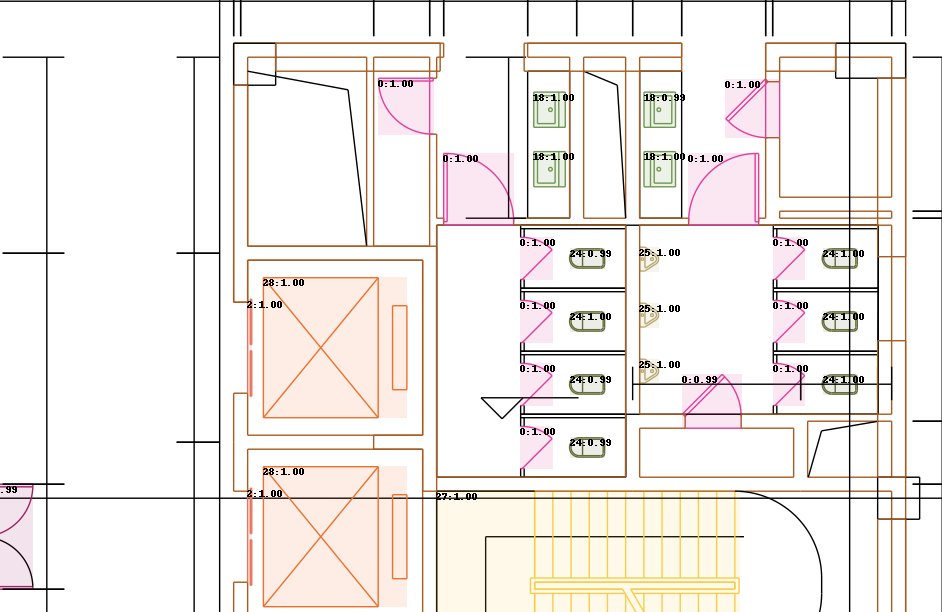}
    \caption{}{Prediction}
    \end{subfigure}
    
\caption{Results of SPv2 on FloorPlanCAD.}
\label{vis}
\end{figure*}

\begin{figure*}[ht!]
    \centering
    \begin{subfigure}{0.45\textwidth}
    \centering
    \includegraphics[width=\textwidth]{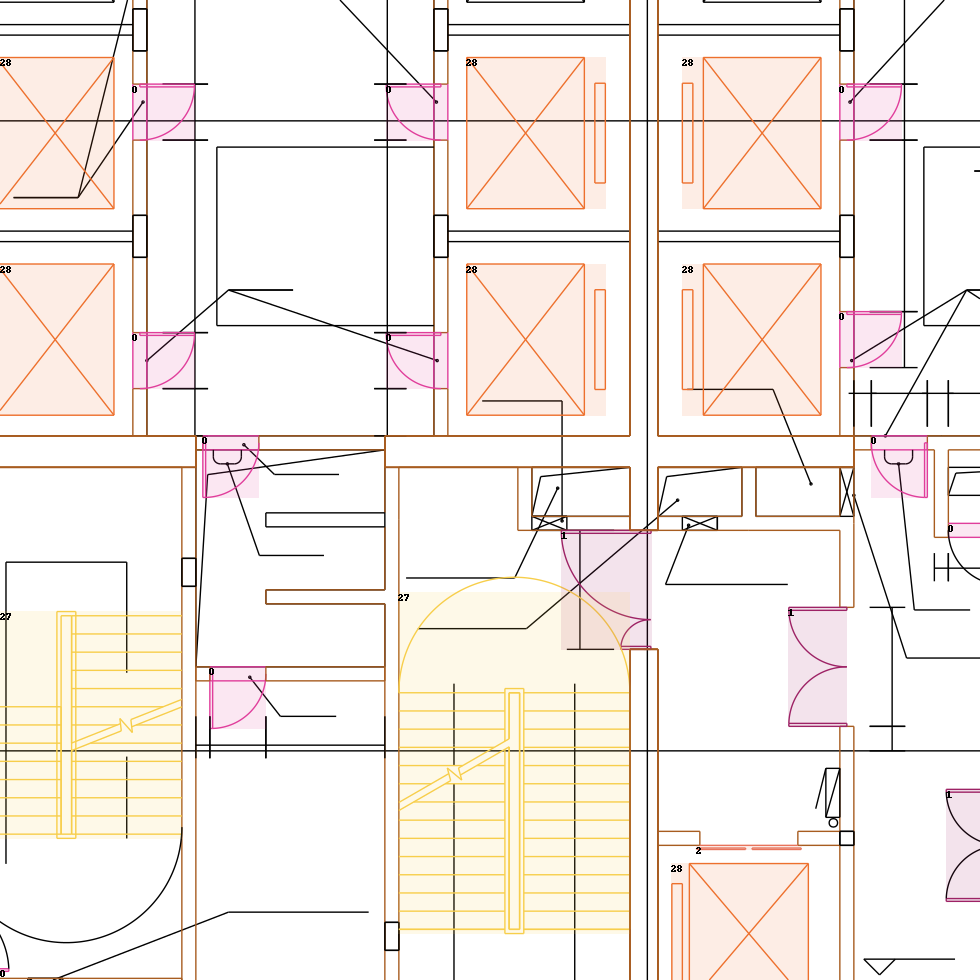}  
    \end{subfigure}
    \begin{subfigure}{0.45\textwidth}
    \centering
    \includegraphics[width=\textwidth]{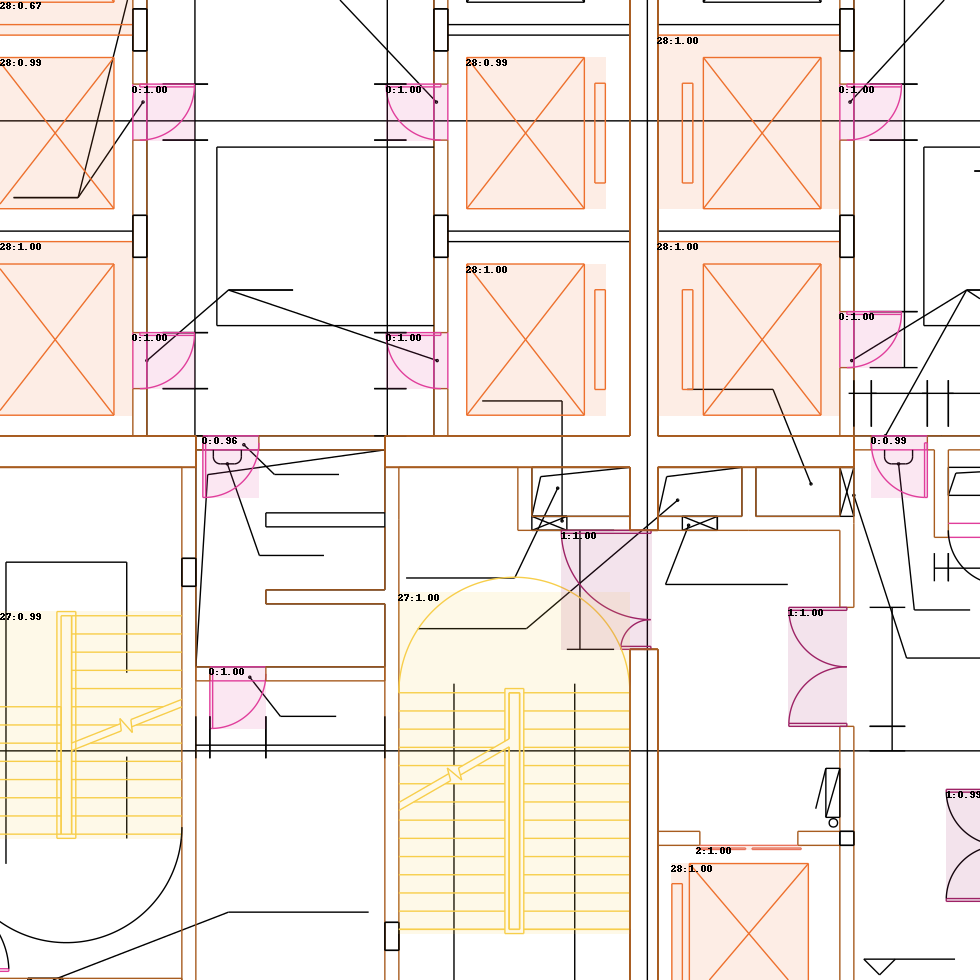}
    \end{subfigure}
    \begin{subfigure}{0.45\textwidth}
    \centering
    \includegraphics[width=\textwidth]{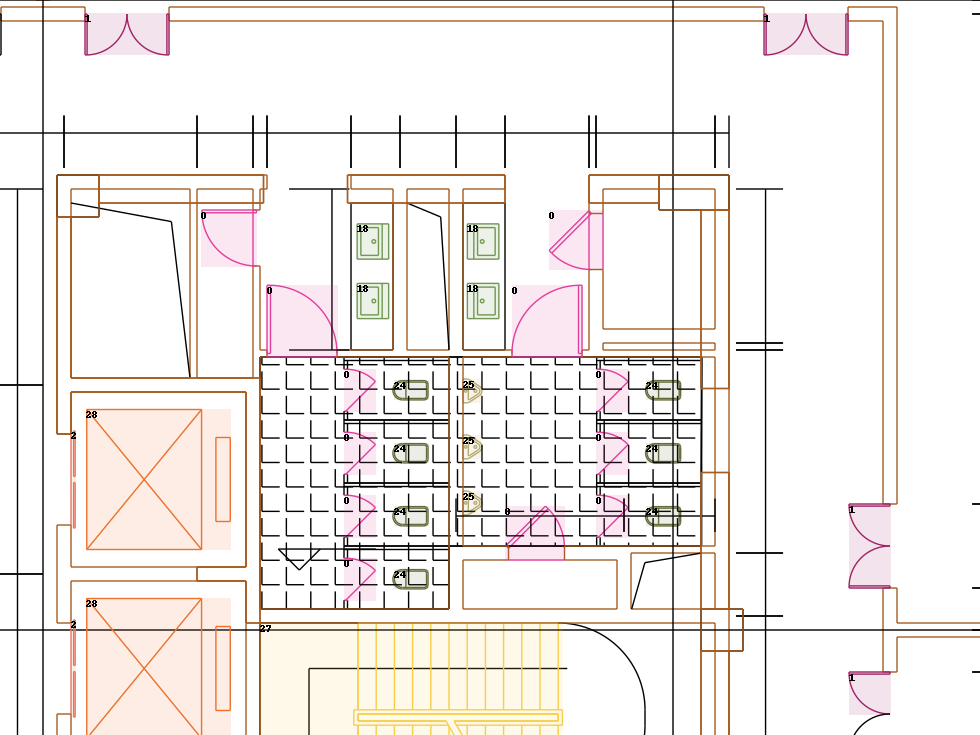}  
    \end{subfigure}
    \begin{subfigure}{0.45\textwidth}
    \centering
    \includegraphics[width=\textwidth]{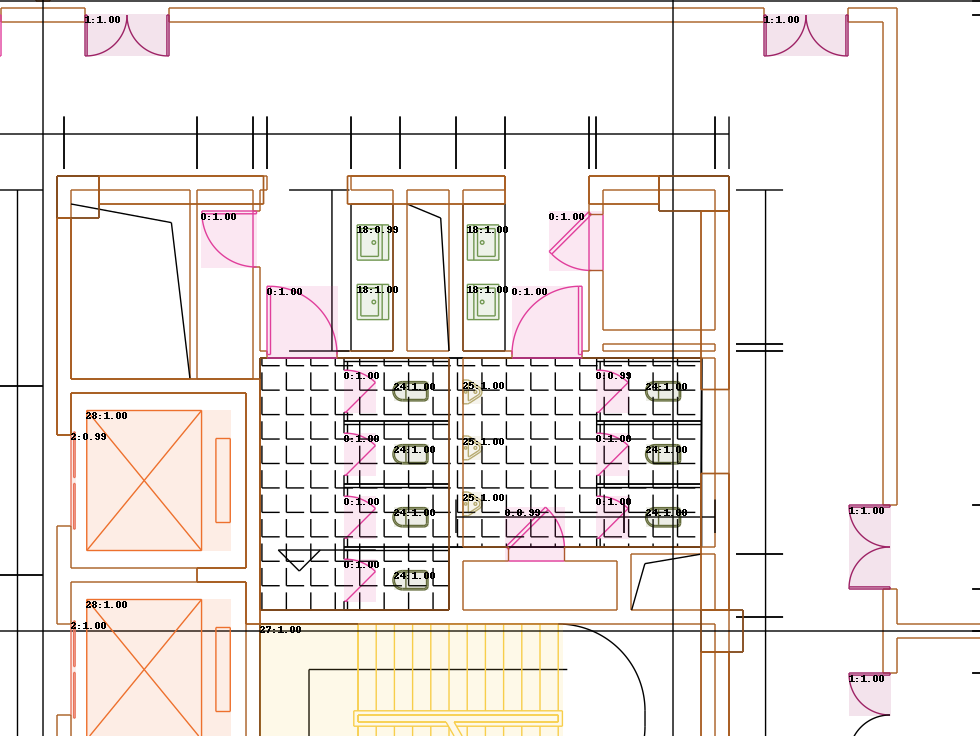}
    \end{subfigure}
    \begin{subfigure}{0.45\textwidth}
    \centering
    \includegraphics[width=\textwidth]{visualization/0910-0013_res_res.png}  
    \caption{}{Ground Truth}
    \end{subfigure}
    \begin{subfigure}{0.45\textwidth}
    \centering
    \includegraphics[width=\textwidth]{visualization/0910-0013_res.png}
    \caption{}{Prediction}
    \end{subfigure}
    
\caption{Results of SPv2 on FloorPlanCAD.}
\label{vis2}
\end{figure*}

\begin{figure*}[ht!]
    \centering
    \begin{subfigure}{0.45\textwidth}
    \centering
    \includegraphics[width=\textwidth]{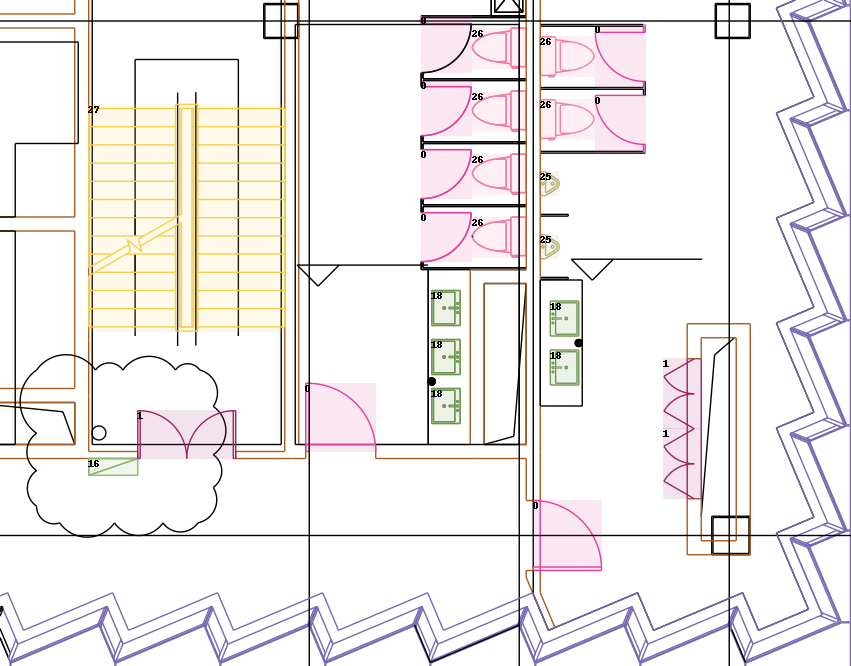}  
    \end{subfigure}
    \begin{subfigure}{0.45\textwidth}
    \centering
    \includegraphics[width=\textwidth]{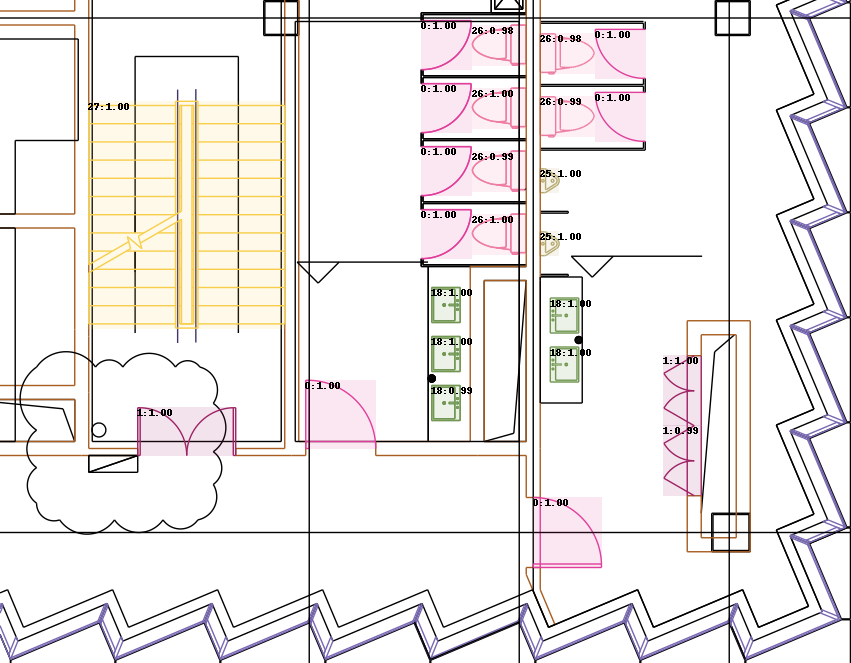}
    \end{subfigure}
    \begin{subfigure}{0.45\textwidth}
    \centering
    \includegraphics[width=\textwidth]{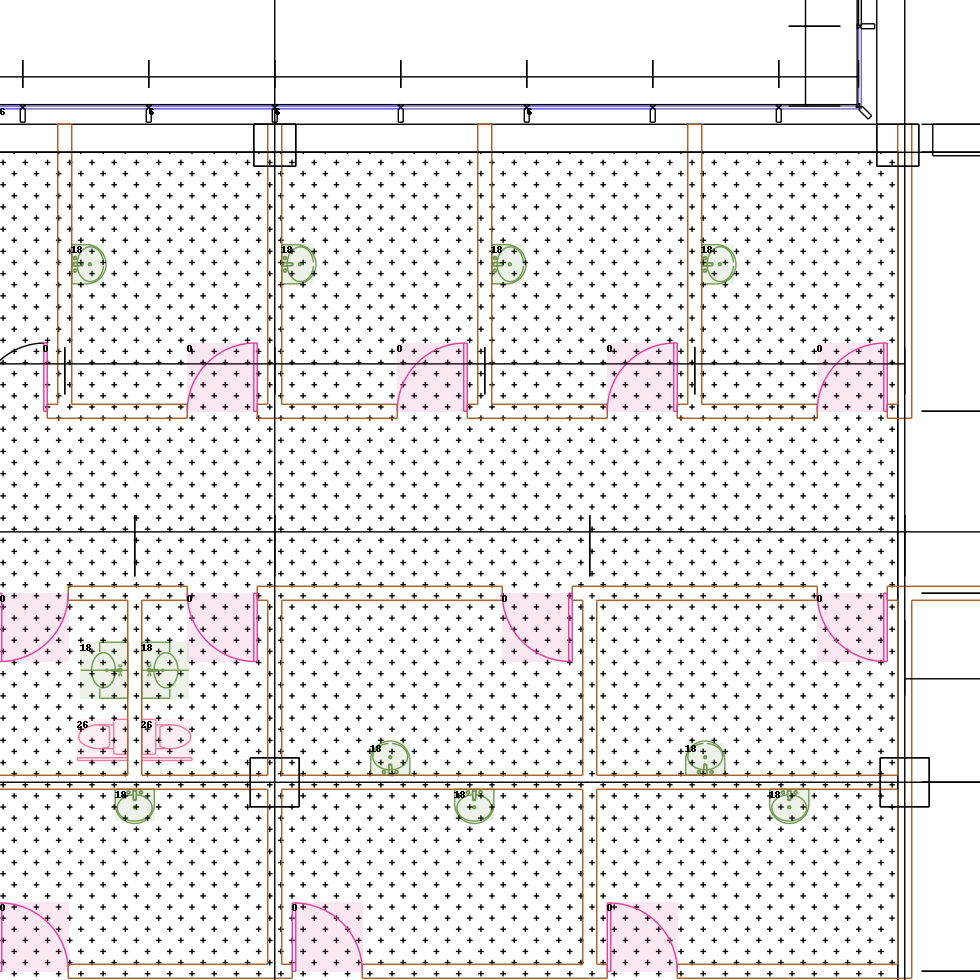}  
    \end{subfigure}
    \begin{subfigure}{0.45\textwidth}
    \centering
    \includegraphics[width=\textwidth]{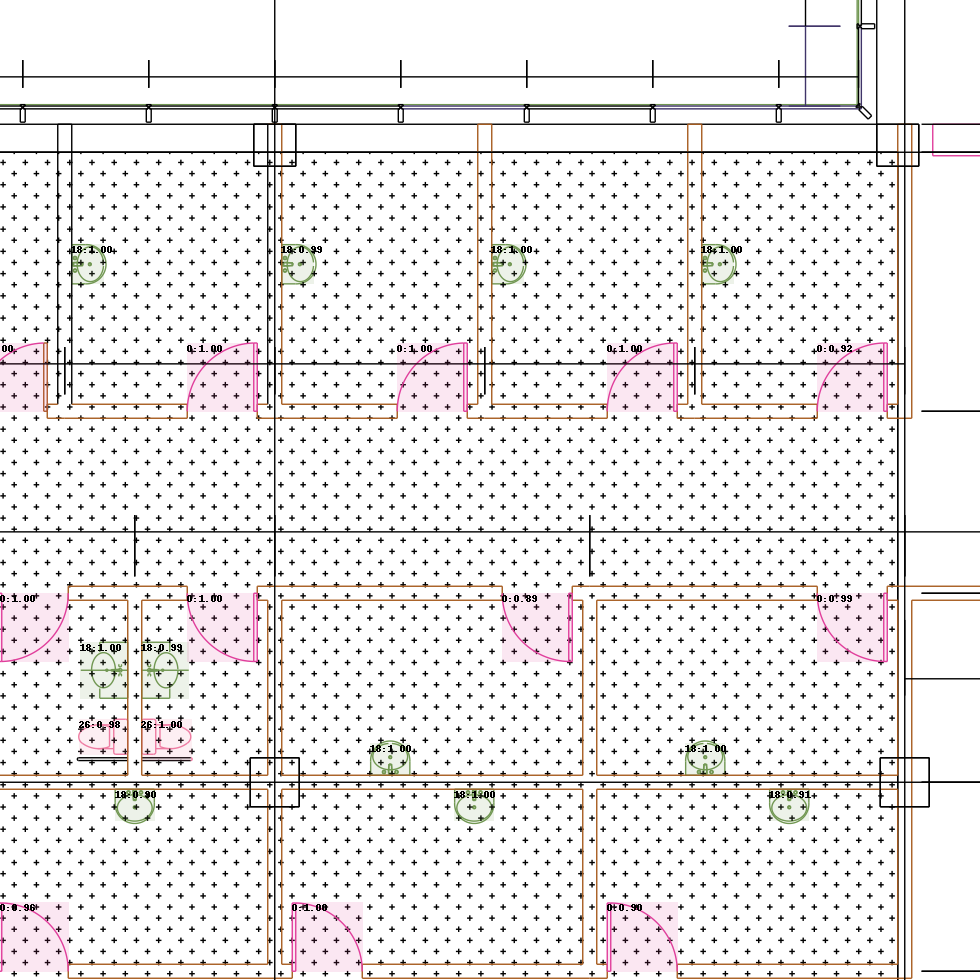}
    \end{subfigure}
    \begin{subfigure}{0.45\textwidth}
    \centering
    \includegraphics[width=\textwidth]{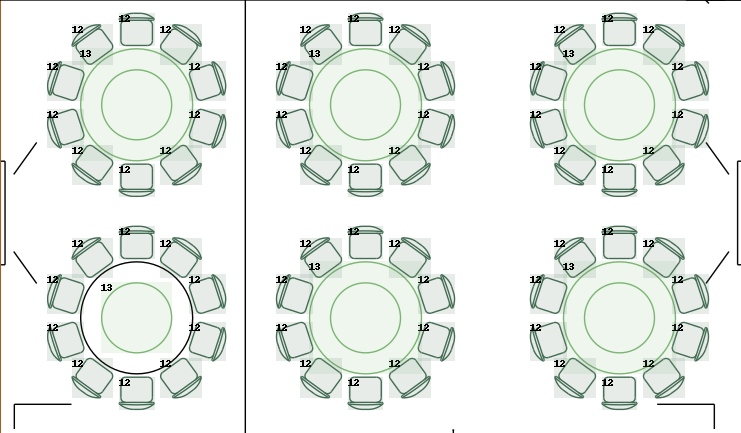}  
    \caption{}{Ground Truth}
    \end{subfigure}
    \begin{subfigure}{0.45\textwidth}
    \centering
    \includegraphics[width=\textwidth]{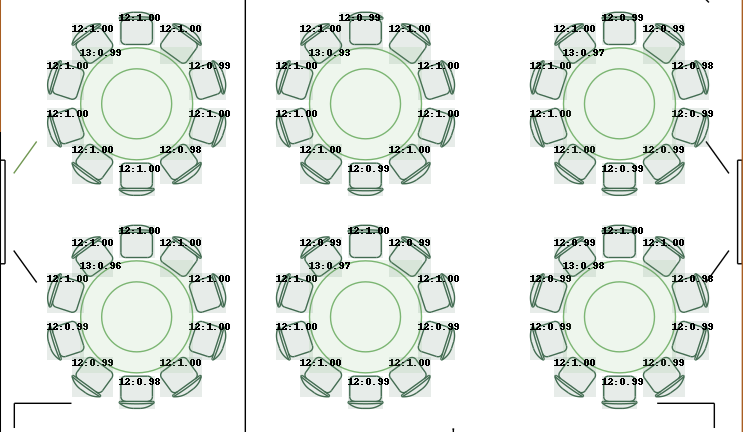}
    \caption{}{Prediction}
    \end{subfigure}
    
\caption{Results of SPv2 on FloorPlanCAD.}
\label{vis3}
\end{figure*}

The results of additional cases are visually represented in this section, you can zoom in on each picture to capture more details, primitives belonging to different classes are represented in distinct colors. More visualized results are shown in Fig.~\ref{vis}~\ref{vis2}~\ref{vis3}.

\end{document}